\documentclass[submission,copyright,creativecommons]{eptcs}

\newcommand{\event}{CAPNS 2018}

\usepackage[english]{babel}
\usepackage{amsmath,amsthm,amssymb} 
\usepackage[svgnames]{xcolor}
\usepackage{rotating}
\usepackage{floatpag}
\usepackage{xifthen} 
\usepackage{bm} 
\usepackage{cite} 
\usepackage{booktabs}
\usepackage{framed}
 
\definecolor{shadecolor}{RGB}{221,160,221}

\usepackage{xspace,enumerate,color,epsfig}   
\usepackage{graphicx}
\graphicspath{{.}{./figures/}}

\usepackage{tikzfig}
\usepackage{stmaryrd}
\usepackage{docmute}
\usepackage{keycommand}

\newkeycommand{\boxmap}[style=small box][1]{\,\tikz{\node[style=\commandkey{style}] (x) {$#1$};\draw(0,-1.25)--(x)--(0,1.25);}\,}
\newkeycommand{\dboxmap}[style=dbox][1]{\,\tikz{\node[style=\commandkey{style}] (x) {$#1$};\draw[boldedge] (0,-1.25)--(x)--(0,1.25);}\,}

\newkeycommand{\wirelabel}[style=white label]{\,\tikz{\node[style=\commandkey{style}]{};}\,\xspace}

\newkeycommand{\pointmap}[style=point][1]{\,\tikz{\node[style=\commandkey{style}] (x) at (0,-0.3) {$#1$};\node [style=none] (3) at (0, 0.9) {};\node [style=none] (3) at (0, -0.9) {};\draw (x)--(0,0.7);}\,}

\newkeycommand{\point}[style=point,estyle=][1]{\,\tikz{\node[style=\commandkey{style}] (x) at (0,-0.05) {$#1$};\draw [\commandkey{estyle}] (x)--(0,1.05);}\,}

\newkeycommand{\copointmap}[style=copoint][1]{\,\tikz{\node[style=\commandkey{style}] (x) at (0,0.3) {$#1$};\draw (x)--(0,-0.7);}\,}

\newkeycommand{\dpoint}[style=point][1]{\,\tikz{\node[style=\commandkey{style},doubled] (x) at (0,-0.3) {$#1$};\draw[boldedge] (x)--(0,0.7);}\,}

\newkeycommand{\kpoint}[style=kpoint,estyle=][1]{\,\tikz{\node[style=\commandkey{style}] (x) at (0,-0.05) {$#1$};\draw [\commandkey{estyle}] (x)--(0,1.05);}\,}

\newkeycommand{\kpointgray}[style=gray kpoint,estyle=][1]{\,\tikz{\node[style=\commandkey{style}] (x) at (0,-0.05) {$#1$};\draw [\commandkey{estyle}] (x)--(0,1.05);}\,}

\newkeycommand{\typedkpoint}[style=kpoint,edgestyle=][2]{%
\,\begin{tikzpicture}
  \begin{pgfonlayer}{nodelayer}
    \node [style=none] (0) at (0, 1) {};
    \node [style=\commandkey{style}] (1) at (0, -0.75) {$#2$};
    \node [style=right label] (2) at (0.25, 0.5) {$#1$};
  \end{pgfonlayer}
  \begin{pgfonlayer}{edgelayer}
    \draw [\commandkey{edgestyle}] (1) to (0.center);
  \end{pgfonlayer}
\end{tikzpicture}\,}

\newkeycommand{\typedkpointdag}[style=kpoint adjoint,edgestyle=][2]{%
\,\begin{tikzpicture}
  \begin{pgfonlayer}{nodelayer}
    \node [style=none] (0) at (0, -1) {};
    \node [style=\commandkey{style}] (1) at (0, 0.75) {$#2$};
    \node [style=right label] (2) at (0.25, -0.5) {$#1$};
  \end{pgfonlayer}
  \begin{pgfonlayer}{edgelayer}
    \draw [\commandkey{edgestyle}] (0.center) to (1);
  \end{pgfonlayer}
\end{tikzpicture}\,}

\newkeycommand{\bistate}[style=kpoint][1]{\,\begin{tikzpicture}
  \begin{pgfonlayer}{nodelayer}
    \node [style=\commandkey{style}, minimum width=1 cm, inner sep=2pt] (0) at (0, -0.25) {$#1$};
    \node [style=none] (1) at (-0.75, 0) {};
    \node [style=none] (2) at (0.75, 0) {};
    \node [style=none] (3) at (-0.75, 0.88) {};
    \node [style=none] (4) at (0.75, 0.88) {};
  \end{pgfonlayer}
  \begin{pgfonlayer}{edgelayer}
    \draw (1.center) to (3.center);
    \draw (2.center) to (4.center);
  \end{pgfonlayer}
\end{tikzpicture}\,}

\newkeycommand{\bistatebig}[style=kpoint][1]{\,\begin{tikzpicture}
  \begin{pgfonlayer}{nodelayer}
    \node [style=\commandkey{style}, minimum width=, minimum width=1.26 cm, inner sep=2pt] (0) at (0, -0.25) {$#1$};
    \node [style=none] (1) at (-1, 0) {};
    \node [style=none] (2) at (1, 0) {};
    \node [style=none] (3) at (-1, 0.88) {};
    \node [style=none] (4) at (1, 0.88) {};
  \end{pgfonlayer}
  \begin{pgfonlayer}{edgelayer}
    \draw (1.center) to (3.center);
    \draw (2.center) to (4.center);
  \end{pgfonlayer}
\end{tikzpicture}\,}

\newkeycommand{\dbistate}[style=dkpoint][1]{\,\begin{tikzpicture}
  \begin{pgfonlayer}{nodelayer}
    \node [style=\commandkey{style}, minimum width=1 cm, inner sep=2pt] (0) at (0, -0.25) {$#1$};
    \node [style=none] (1) at (-0.75, 0) {};
    \node [style=none] (2) at (0.75, 0) {};
    \node [style=none] (3) at (-0.75, 1.0) {};
    \node [style=none] (4) at (0.75, 1.0) {};
  \end{pgfonlayer}
  \begin{pgfonlayer}{edgelayer}
    \draw [boldedge] (1.center) to (3.center);
    \draw [boldedge] (2.center) to (4.center);
  \end{pgfonlayer}
\end{tikzpicture}\,}

\newkeycommand{\bistatebraket}[style1=kpoint,style2=kpointadj][2]{\,\begin{tikzpicture}
  \begin{pgfonlayer}{nodelayer}
    \node [style=\commandkey{style1}, minimum width=1 cm, inner sep=2pt] (0) at (0, -0.75) {$#1$};
    \node [style=none] (1) at (-0.75, -0.5) {};
    \node [style=none] (2) at (0.75, -0.5) {};
    \node [style=none] (3) at (-0.75, 0.5) {};
    \node [style=none] (4) at (0.75, 0.5) {};
    \node [style=\commandkey{style2}, minimum width=1 cm, inner sep=2pt] (5) at (0, 0.75) {$#2$};
  \end{pgfonlayer}
  \begin{pgfonlayer}{edgelayer}
    \draw (1.center) to (3.center);
    \draw (2.center) to (4.center);
  \end{pgfonlayer}
\end{tikzpicture}\,}

\newkeycommand{\weight}[style=dkpoint][1]{\,\begin{tikzpicture}
  \begin{pgfonlayer}{nodelayer}
    \node [style=\commandkey{style}] (0) at (0, -0.5) {$#1$};
    \node [style=upground] (1) at (0, 0.75) {};
  \end{pgfonlayer}
  \begin{pgfonlayer}{edgelayer}
    \draw [style=boldedge] (0) to (1);
  \end{pgfonlayer}
\end{tikzpicture}\,}

\newkeycommand{\dkpoint}[style=dkpoint][1]{\,\tikz{\node[style=\commandkey{style}] (x) at (0,-0.1) {$#1$};\draw[boldedge] (x)--(0,1);}\,}
\newkeycommand{\dkpointadj}[style=dkpointadj][1]{\,\tikz{\node[style=\commandkey{style}] (x) at (0,0.1) {$#1$};\draw[boldedge] (x)--(0,-1);}\,}
\newkeycommand{\dkpointtrans}[style=dkpointtrans][1]{\,\tikz{\node[style=\commandkey{style}] (x) at (0,0.1) {$#1$};\draw[boldedge] (x)--(0,-1);}\,}
\newkeycommand{\dkpointconj}[style=dkpointconj][1]{\,\tikz{\node[style=\commandkey{style}] (x) at (0,-0.1) {$#1$};\draw[boldedge] (x)--(0,1);}\,}

\newkeycommand{\blackkpoint}[style=black kpoint][1]{\,\tikz{\node[style=\commandkey{style}] (x) at (0,-0.1) {$#1$};\draw (x)--(0,1);}\,}
\newkeycommand{\blackkpointadj}[style=black kpointadj][1]{\,\tikz{\node[style=\commandkey{style}] (x) at (0,0.1) {$#1$};\draw (x)--(0,-1);}\,}
\newkeycommand{\blackdkpoint}[style=black dkpoint][1]{\,\tikz{\node[style=\commandkey{style}] (x) at (0,-0.1) {$#1$};\draw[boldedge] (x)--(0,1);}\,}
\newkeycommand{\blackdkpointadj}[style=black dkpointadj][1]{\,\tikz{\node[style=\commandkey{style}] (x) at (0,0.1) {$#1$};\draw[boldedge] (x)--(0,-1);}\,}

\newcommand{\trace}{\,\begin{tikzpicture}
  \begin{pgfonlayer}{nodelayer}
    \node [style=none] (0) at (0, -0.60) {};
    \node [style=upground] (1) at (0, 0.40) {};
  \end{pgfonlayer}
  \begin{pgfonlayer}{edgelayer}
    \draw [style=boldedge] (0.center) to (1);
  \end{pgfonlayer}
\end{tikzpicture}\,}

\newcommand\discard\trace

\newkeycommand{\pointketbra}[style1=copoint,style2=point,estyle=][2]{\,%
\begin{tikzpicture}
  \begin{pgfonlayer}{nodelayer}
    \node [style=none] (0) at (0, -1.75) {};
    \node [style=\commandkey{style1}] (1) at (0, -0.75) {$#1$};
    \node [style=\commandkey{style2}] (2) at (0, 0.75) {$#2$};
    \node [style=none] (3) at (0, 1.75) {};
  \end{pgfonlayer}
  \begin{pgfonlayer}{edgelayer}
    \draw [\commandkey{estyle}] (0.center) to (1);
    \draw [\commandkey{estyle}] (2) to (3.center);
  \end{pgfonlayer}
\end{tikzpicture}\,}

\newkeycommand{\twocopointketbra}[style1=copoint,style2=copoint,style3=point][3]{\,%
\begin{tikzpicture}
  \begin{pgfonlayer}{nodelayer}
    \node [style=none] (0) at (-0.75, -1.75) {};
    \node [style=none] (1) at (0.75, -1.75) {};
    \node [style=\commandkey{style1}] (2) at (-0.75, -0.75) {$#1$};
    \node [style=\commandkey{style2}] (3) at (0.75, -0.75) {$#2$};
    \node [style=\commandkey{style3}] (4) at (0, 0.75) {$#3$};
    \node [style=none] (5) at (0, 1.75) {};
  \end{pgfonlayer}
  \begin{pgfonlayer}{edgelayer}
    \draw (0.center) to (2);
    \draw (1.center) to (3);
    \draw (4) to (5.center);
  \end{pgfonlayer}
\end{tikzpicture}\,}

\newkeycommand{\twopointketbra}[style1=copoint,style2=point,style3=point][3]{\,%
\begin{tikzpicture}
  \begin{pgfonlayer}{nodelayer}
    \node [style=none] (0) at (0, -1.75) {};
    \node [style=none] (1) at (-0.75, 1.75) {};
    \node [style=\commandkey{style1}] (2) at (0, -0.75) {$#1$};
    \node [style=\commandkey{style2}] (3) at (-0.75, 0.75) {$#2$};
    \node [style=\commandkey{style3}] (4) at (0.75, 0.75) {$#3$};
    \node [style=none] (5) at (0.75, 1.75) {};
  \end{pgfonlayer}
  \begin{pgfonlayer}{edgelayer}
    \draw (0.center) to (2);
    \draw (1.center) to (3);
    \draw (4) to (5.center);
  \end{pgfonlayer}
\end{tikzpicture}\,}

\newkeycommand{\pointbraket}[style1=point,style2=copoint,estyle=][2]{\,%
\begin{tikzpicture}
  \begin{pgfonlayer}{nodelayer}
    \node [style=\commandkey{style1}] (0) at (0, -0.625) {$#1$};
    \node [style=\commandkey{style2}] (1) at (0, 0.625) {$#2$};
  \end{pgfonlayer}
  \begin{pgfonlayer}{edgelayer}
    \draw [\commandkey{estyle}] (0) to (1);
  \end{pgfonlayer}
\end{tikzpicture}\,}

\newkeycommand{\kpointketbra}[style1=kpointdag,style2=kpoint,estyle=][2]{\,%
\begin{tikzpicture}
  \begin{pgfonlayer}{nodelayer}
    \node [style=none] (0) at (0, -1.9) {};
    \node [style=\commandkey{style1}] (1) at (0, -0.95) {$#1$};
    \node [style=\commandkey{style2}] (2) at (0, 0.95) {$#2$};
    \node [style=none] (3) at (0, 1.9) {};
  \end{pgfonlayer}
  \begin{pgfonlayer}{edgelayer}
    \draw [\commandkey{estyle}] (0.center) to (1);
    \draw [\commandkey{estyle}] (2) to (3.center);
  \end{pgfonlayer}
\end{tikzpicture}\,}

\newkeycommand{\kpointmap}[style1=kpoint,style2=map][2]{\,%
\begin{tikzpicture}
  \begin{pgfonlayer}{nodelayer}
    \node [style=\commandkey{style1}] (0) at (0, -1.1) {$#1$};
    \node [style=\commandkey{style2}] (1) at (0, 0.5) {$#2$};
    \node [style=none] (2) at (0, 1.75) {};
  \end{pgfonlayer}
  \begin{pgfonlayer}{edgelayer}
    \draw (0) to (1);
    \draw (1) to (2.center);
  \end{pgfonlayer}
\end{tikzpicture}%
\,}

\newkeycommand{\kpointmaptrans}[style1=kpointconj,style2=mapadj][2]{\,%
\begin{tikzpicture}
  \begin{pgfonlayer}{nodelayer}
    \node [style=\commandkey{style1}] (0) at (0, -1.1) {$#1$};
    \node [style=\commandkey{style2}] (1) at (0, 0.5) {$#2$};
    \node [style=none] (2) at (0, 1.75) {};
  \end{pgfonlayer}
  \begin{pgfonlayer}{edgelayer}
    \draw (0) to (1);
    \draw (1) to (2.center);
  \end{pgfonlayer}
\end{tikzpicture}%
\,}

\newkeycommand{\kpointmapadj}[style1=kpointadj,style2=map][2]{\,%
\begin{tikzpicture}
  \begin{pgfonlayer}{nodelayer}
    \node [style=\commandkey{style1}] (0) at (0, 1.1) {$#1$};
    \node [style=\commandkey{style2}] (1) at (0, -0.5) {$#2$};
    \node [style=none] (2) at (0, -1.75) {};
  \end{pgfonlayer}
  \begin{pgfonlayer}{edgelayer}
    \draw (0) to (1);
    \draw (1) to (2.center);
  \end{pgfonlayer}
\end{tikzpicture}%
\,}

\newkeycommand{\dkpointmap}[style1=dkpoint,style2=dmap][2]{\,%
\begin{tikzpicture}
  \begin{pgfonlayer}{nodelayer}
    \node [style=\commandkey{style1}] (0) at (0, -1.2) {$#1$};
    \node [style=\commandkey{style2}] (1) at (0, 0.4) {$#2$};
    \node [style=none] (2) at (0, 1.7) {};
  \end{pgfonlayer}
  \begin{pgfonlayer}{edgelayer}
    \draw[style=boldedge] (0) to (1);
    \draw[style=boldedge] (1) to (2.center);
  \end{pgfonlayer}
\end{tikzpicture}%
\,}

\newkeycommand{\dkpointmapx}[style1=dkpoint,style2=dmap][2]{\,%
\begin{tikzpicture}
  \begin{pgfonlayer}{nodelayer}
    \node [style=\commandkey{style1}] (0) at (0, -1.4) {$#1$};
    \node [style=\commandkey{style2}] (1) at (0, 0.6) {$#2$};
    \node [style=none] (2) at (0, 1.9) {};
  \end{pgfonlayer}
  \begin{pgfonlayer}{edgelayer}
    \draw[style=boldedge] (0) to (1);
    \draw[style=boldedge] (1) to (2.center);
  \end{pgfonlayer}
\end{tikzpicture}%
\,}

\newkeycommand{\boxcopointmapdag}[style1=map,style2=copoint][2]{\,%
\begin{tikzpicture}
  \begin{pgfonlayer}{nodelayer}
    \node [style=none] (0) at (0, -1.75) {};
    \node [style=\commandkey{style1}] (1) at (0, -0.5) {$#1$};
    \node [style=\commandkey{style2}] (2) at (0, 1) {$#2$};
  \end{pgfonlayer}
  \begin{pgfonlayer}{edgelayer}
    \draw (0.center) to (1);
    \draw (1) to (2);
  \end{pgfonlayer}
\end{tikzpicture}%
\,}

\newkeycommand{\kpointdagmap}[style1=map,style2=kpointdag][2]{\,%
\begin{tikzpicture}
  \begin{pgfonlayer}{nodelayer}
    \node [style=none] (0) at (0, -1.75) {};
    \node [style=\commandkey{style1}] (1) at (0, -0.5) {$#1$};
    \node [style=\commandkey{style2}] (2) at (0, 1.1) {$#2$};
  \end{pgfonlayer}
  \begin{pgfonlayer}{edgelayer}
    \draw (0.center) to (1);
    \draw (1) to (2);
  \end{pgfonlayer}
\end{tikzpicture}%
\,}

\newkeycommand{\sandwichmap}[style1=point,style2=map,style3=copoint][3]{\,%
\begin{tikzpicture}
  \begin{pgfonlayer}{nodelayer}
    \node [style=\commandkey{style1}] (0) at (0, -1.50) {$#1$};
    \node [style=\commandkey{style2}] (1) at (0, 0) {$#2$};
    \node [style=\commandkey{style3}] (2) at (0, 1.50) {$#3$};
  \end{pgfonlayer}
  \begin{pgfonlayer}{edgelayer}
    \draw (0) to (1);
    \draw (1) to (2);
  \end{pgfonlayer}
\end{tikzpicture}%
}

\newkeycommand{\kpointsandwichmap}[style1=kpoint,style2=map,style3=kpointdag][3]{\,%
\begin{tikzpicture}
  \begin{pgfonlayer}{nodelayer}
    \node [style=\commandkey{style1}] (0) at (0, -1.5) {$#1$};
    \node [style=\commandkey{style2}] (1) at (0, 0) {$#2$};
    \node [style=\commandkey{style3}] (2) at (0, 1.5) {$#3$};
  \end{pgfonlayer}
  \begin{pgfonlayer}{edgelayer}
    \draw (0) to (1);
    \draw (1) to (2);
  \end{pgfonlayer}
\end{tikzpicture}%
}

\newkeycommand{\sandwichmapconj}[style1=point,style2=mapconj,style3=copoint][3]{\,%
\begin{tikzpicture}
  \begin{pgfonlayer}{nodelayer}
    \node [style=\commandkey{style1}] (0) at (0, -1.50) {$#1$};
    \node [style=\commandkey{style2}] (1) at (0, 0) {$#2$};
    \node [style=\commandkey{style3}] (2) at (0, 1.50) {$#3$};
  \end{pgfonlayer}
  \begin{pgfonlayer}{edgelayer}
    \draw (0) to (1);
    \draw (1) to (2);
  \end{pgfonlayer}
\end{tikzpicture}%
}

\newkeycommand{\sandwichtwo}[style1=point,style2=map,style3=map,style4=copoint][4]{\,%
\begin{tikzpicture}
  \begin{pgfonlayer}{nodelayer}
    \node [style=\commandkey{style2}] (0) at (0, -1) {$#2$};
    \node [style=\commandkey{style3}] (1) at (0, 1) {$#3$};
    \node [style=\commandkey{style1}] (2) at (0, -2.6) {$#1$};
    \node [style=\commandkey{style4}] (3) at (0, 2.6) {$#4$};
  \end{pgfonlayer}
  \begin{pgfonlayer}{edgelayer}
    \draw (2) to (0);
    \draw (0) to (1);
    \draw (1) to (3);
  \end{pgfonlayer}
\end{tikzpicture}\,}

\newkeycommand{\longbraket}[style1=point,style2=copoint][2]{\,%
\begin{tikzpicture}
  \begin{pgfonlayer}{nodelayer}
    \node [style=\commandkey{style1}] (0) at (0, -2.6) {$#1$};
    \node [style=\commandkey{style2}] (1) at (0, 2.6) {$#2$};
  \end{pgfonlayer}
  \begin{pgfonlayer}{edgelayer}
    \draw (0) to (1);
  \end{pgfonlayer}
\end{tikzpicture}\,}

\newkeycommand{\onb}[style=point]{\ensuremath{\{\,\tikz{\node[style=\commandkey{style}] (x) at (0,-0.3) {$j$};\draw (x)--(0,0.7);}\,\}}\xspace}

\newkeycommand{\phase}[style=white phase dot][1]{\,\begin{tikzpicture}
    \begin{pgfonlayer}{nodelayer}
        \node [style=none] (0) at (0, 1) {};
        \node [style=\commandkey{style}] (2) at (0, -0) {$#1$}; 
        \node [style=none] (3) at (0, -1) {};
    \end{pgfonlayer}
    \begin{pgfonlayer}{edgelayer}
        \draw (2) to (0.center);
        \draw (3.center) to (2);
    \end{pgfonlayer}
\end{tikzpicture}\,}

\newkeycommand{\dphase}[style=white phase ddot][1]{\,\begin{tikzpicture}
    \begin{pgfonlayer}{nodelayer}
        \node [style=none] (0) at (0, 1.25) {};
        \node [style=\commandkey{style}] (2) at (0, -0) {$#1$};
        \node [style=none] (3) at (0, -1.25) {};
    \end{pgfonlayer}
    \begin{pgfonlayer}{edgelayer}
        \draw [boldedge] (2) to (0.center);
        \draw [boldedge] (3.center) to (2);
    \end{pgfonlayer}
\end{tikzpicture}\,}

\newkeycommand{\dphasepoint}[style=white phase ddot][1]{\,\begin{tikzpicture}[yshift=-3mm]
  \begin{pgfonlayer}{nodelayer}
    \node [style=none] (0) at (0, 1) {};
    \node [style=\commandkey{style}] (1) at (0, 0) {$#1$};
  \end{pgfonlayer}
  \begin{pgfonlayer}{edgelayer}
    \draw [style=boldedge] (0.center) to (1);
  \end{pgfonlayer}
\end{tikzpicture}\,}

\newkeycommand{\dphasepointgray}[style=gray phase ddot][1]{\,\begin{tikzpicture}[yshift=-3mm]
  \begin{pgfonlayer}{nodelayer}
    \node [style=none] (0) at (0, 1) {};
    \node [style=\commandkey{style}] (1) at (0, 0) {$#1$};
  \end{pgfonlayer}
  \begin{pgfonlayer}{edgelayer}
    \draw [style=boldedge] (0.center) to (1);
  \end{pgfonlayer}
\end{tikzpicture}\,}

\newkeycommand{\dphasecopoint}[style=white phase ddot][1]{\,\begin{tikzpicture}[yshift=3mm,yscale=-1]
  \begin{pgfonlayer}{nodelayer}
    \node [style=none] (0) at (0, 1) {};
    \node [style=\commandkey{style}] (1) at (0, 0) {$#1$};
  \end{pgfonlayer}
  \begin{pgfonlayer}{edgelayer}
    \draw [style=boldedge] (0.center) to (1);
  \end{pgfonlayer}
\end{tikzpicture}\,}

\newkeycommand{\dphasepointsm}[style=white phase ddot][1]{\,\begin{tikzpicture}[yshift=-3mm]
  \begin{pgfonlayer}{nodelayer}
    \node [style=none] (0) at (0, 1) {};
    \node [style=\commandkey{style}] (1) at (0, 0) {\footnotesize\!$#1$\!};
  \end{pgfonlayer}
  \begin{pgfonlayer}{edgelayer}
    \draw [style=boldedge] (0.center) to (1);
  \end{pgfonlayer}
\end{tikzpicture}\,}

\newkeycommand{\phasepoint}[style=white phase dot][1]{\,\begin{tikzpicture}[yshift=-3mm]
  \begin{pgfonlayer}{nodelayer}
    \node [style=none] (0) at (0, 1) {};
    \node [style=\commandkey{style}] (1) at (0, 0) {$#1$};
  \end{pgfonlayer}
  \begin{pgfonlayer}{edgelayer}
    \draw (0.center) to (1);
  \end{pgfonlayer}
\end{tikzpicture}\,}

\newkeycommand{\phasecopoint}[style=white phase dot][1]{\,\begin{tikzpicture}[yshift=3mm]
  \begin{pgfonlayer}{nodelayer}
    \node [style=none] (0) at (0, -1) {};
    \node [style=\commandkey{style}] (1) at (0, 0) {$#1$};
  \end{pgfonlayer}
  \begin{pgfonlayer}{edgelayer}
    \draw (0.center) to (1);
  \end{pgfonlayer}
\end{tikzpicture}\,}

\newkeycommand{\kpointbraket}[style1=kpoint,style2=kpoint adjoint,estyle=][2]{\,%
\begin{tikzpicture}
  \begin{pgfonlayer}{nodelayer}
    \node [style=\commandkey{style1}] (0) at (0, -0.735) {$#1$};
    \node [style=\commandkey{style2}] (1) at (0, 0.735) {$#2$};
  \end{pgfonlayer}
  \begin{pgfonlayer}{edgelayer}
    \draw [\commandkey{estyle}] (0) to (1);
  \end{pgfonlayer}
\end{tikzpicture}\,}

\newkeycommand{\kkpointbraket}[style1=kpoint,style2=kpoint adjoint,estyle=][2]{\,%
\begin{tikzpicture}
  \begin{pgfonlayer}{nodelayer}
    \node [style=\commandkey{style1}] (0) at (0, -0.685) {$#1$};
    \node [style=\commandkey{style2}] (1) at (0, 0.685) {$#2$};
  \end{pgfonlayer}
  \begin{pgfonlayer}{edgelayer}
    \draw [\commandkey{estyle}] (0) to (1);
  \end{pgfonlayer}
\end{tikzpicture}\,}

\newkeycommand{\kpointbraketx}[style1=kpoint,style2=kpoint adjoint,estyle=][2]{\,%
\begin{tikzpicture}
  \begin{pgfonlayer}{nodelayer}
    \node [style=\commandkey{style1}] (0) at (0, -0.885) {$#1$};
    \node [style=\commandkey{style2}] (1) at (0, 0.885) {$#2$};
  \end{pgfonlayer}
  \begin{pgfonlayer}{edgelayer}
    \draw [\commandkey{estyle}] (0) to (1);
  \end{pgfonlayer}
\end{tikzpicture}\,}

\tikzstyle{cdiag}=[matrix of math nodes, row sep=3em, column sep=3em, text height=1.5ex, text depth=0.25ex,inner sep=0.5em]
\tikzstyle{arrow above}=[transform canvas={yshift=0.5ex}]
\tikzstyle{arrow below}=[transform canvas={yshift=-0.5ex}]

\usepackage[svgnames]{xcolor}
\usepackage{tikz}
\usetikzlibrary{decorations.markings}
\usetikzlibrary{shapes.geometric}
\pgfdeclarelayer{edgelayer}
\pgfdeclarelayer{nodelayer}
\pgfsetlayers{edgelayer,nodelayer,main}

\tikzstyle{none}=[inner sep=0pt]
\definecolor{hexcolor0xff0000}{rgb}{1.000,0.000,0.000}
\definecolor{hexcolor0x000000}{rgb}{0.000,0.000,0.000}
\definecolor{hexcolor0x00ff00}{rgb}{0.000,1.000,0.000}
\definecolor{hexcolor0x000000}{rgb}{0.000,0.000,0.000}
\definecolor{hexcolor0xffff00}{rgb}{1.000,1.000,0.000}
\definecolor{hexcolor0xffffff}{rgb}{1.000,1.000,1.000}

\tikzstyle{rn}=[circle,fill=hexcolor0xff0000,draw=hexcolor0x000000,line width=0.8 pt]
\tikzstyle{gn}=[circle,fill=hexcolor0x00ff00,draw=hexcolor0x000000,line width=0.8 pt]
\tikzstyle{yn}=[circle,fill=hexcolor0xffff00,draw=hexcolor0x000000,line width=0.8 pt]
\tikzstyle{wn}=[circle,fill=hexcolor0xffffff,draw=hexcolor0x000000,line width=0.8 pt]
\tikzstyle{wnthick}=[circle,fill=hexcolor0xffffff,draw=hexcolor0x000000,line width=2.500]

\tikzstyle{simple}=[-,draw=hexcolor0x000000,line width=2.000]
\tikzstyle{arrow}=[-,draw=hexcolor0x000000,postaction={decorate},decoration={markings,mark=at position .5 with {\arrow{>}}},line width=2.000]
\tikzstyle{tick}=[-,draw=hexcolor0x000000,postaction={decorate},decoration={markings,mark=at position .5 with {\draw (0,-0.1) -- (0,0.1);}},line width=2.000]
\tikzstyle{halfthickness}=[-,draw=hexcolor0x000000,line width=0.500]
\tikzstyle{thick}=[-,draw=hexcolor0x000000,line width=2.500]
\tikzstyle{thicker}=[-,draw=hexcolor0x000000,line width=4.000]

\tikzstyle{env}=[copoint,regular polygon rotate=0,minimum width=0.2cm, fill=black]

\tikzstyle{probs}=[shape=semicircle,fill=white,draw=black,shape border rotate=180,minimum width=1.2cm]

\tikzstyle{every picture}=[baseline=-0.25em,scale=0.5]
\tikzstyle{dotpic}=[] 
\tikzstyle{diredges}=[every to/.style={diredge}]
\tikzstyle{math matrix}=[matrix of math nodes,left delimiter=(,right delimiter=),inner sep=2pt,column sep=1em,row sep=0.5em,nodes={inner sep=0pt},text height=1.5ex, text depth=0.25ex]

\tikzstyle{inline text}=[text height=1.5ex, text depth=0.25ex,yshift=0.5mm]
\tikzstyle{label}=[font=\footnotesize,text height=1.5ex, text depth=0.25ex,yshift=0.5mm]
\tikzstyle{left label}=[label,anchor=east,xshift=1.5mm]
\tikzstyle{right label}=[label,anchor=west,xshift=-1.5mm]

\tikzstyle{braceedge}=[decorate,decoration={brace,amplitude=2mm,raise=-1mm}]
\tikzstyle{small braceedge}=[decorate,decoration={brace,amplitude=1mm,raise=-1mm}]

\tikzstyle{doubled}=[line width=1.6pt] 
\tikzstyle{boldedge}=[doubled,shorten <=-0.17mm,shorten >=-0.17mm]
\tikzstyle{boldedgegray}=[doubled,gray,shorten <=-0.17mm,shorten >=-0.17mm]

\tikzstyle{semidoubled}=[line width=1.4pt] 
\tikzstyle{semiboldedgegray}=[semidoubled,gray,shorten <=-0.17mm,shorten >=-0.17mm]

\tikzstyle{boldedgedashed}=[very thick,dashed,shorten <=-0.17mm,shorten >=-0.17mm]
\tikzstyle{vboldedgedashed}=[doubled,dashed,shorten <=-0.17mm,shorten >=-0.17mm]
\tikzstyle{left hook arrow}=[left hook-latex]
\tikzstyle{right hook arrow}=[right hook-latex]
\tikzstyle{sembracket}=[line width=0.5pt,shorten <=-0.07mm,shorten >=-0.07mm]

\tikzstyle{causal edge}=[->,thick,gray]
\tikzstyle{causal nondir}=[thick,gray]
\tikzstyle{timeline}=[thick,gray, dashed]

\tikzstyle{cedge}=[<->,thick,gray!70!white]

\tikzstyle{empty diagram}=[draw=gray!40!white,dashed,shape=rectangle,minimum width=1cm,minimum height=1cm]
\tikzstyle{empty diagram small}=[draw=gray!50!white,dashed,shape=rectangle,minimum width=0.6cm,minimum height=0.5cm]

\tikzstyle{dot}=[inner sep=0mm,minimum width=2mm,minimum height=2mm,draw,shape=circle]
\tikzstyle{ddot}=[inner sep=0mm, doubled, minimum width=2.5mm,minimum height=2.5mm,draw,shape=circle]

\tikzstyle{black dot}=[dot,fill=black]
\tikzstyle{white dot}=[dot,fill=white,,text depth=-0.2mm]
\tikzstyle{green dot}=[white dot] 
\tikzstyle{gray dot}=[dot,fill=gray!40!white,,text depth=-0.2mm]
\tikzstyle{red dot}=[gray dot] 

\tikzstyle{black ddot}=[ddot,fill=black]
\tikzstyle{white ddot}=[ddot,fill=white]
\tikzstyle{gray ddot}=[ddot,fill=gray!40!white]

\tikzstyle{gray edge}=[gray!40!white]

\tikzstyle{small dot}=[inner sep=0.5mm,minimum width=0pt,minimum height=0pt,draw,shape=circle]

\tikzstyle{small black dot}=[small dot,fill=black]
\tikzstyle{small white dot}=[small dot,fill=white]
\tikzstyle{small gray dot}=[small dot,fill=gray!40!white]

\tikzstyle{causal dot}=[inner sep=0.4mm,minimum width=0pt,minimum height=0pt,draw=white,shape=circle,fill=gray!40!white]

\tikzstyle{phase dimensions}=[minimum size=5mm,font=\footnotesize,rectangle,rounded corners=2.5mm,inner sep=0.2mm,outer sep=-2mm]
\tikzstyle{dphase dimensions}=[minimum size=5mm,font=\footnotesize,rectangle,rounded corners=2.5mm,inner sep=0.2mm,outer sep=-2mm]

\tikzstyle{white phase dot}=[dot,fill=white,phase dimensions]
\tikzstyle{white phase ddot}=[ddot,fill=white,dphase dimensions]

\tikzstyle{white rect ddot}=[draw=black,fill=white,doubled,minimum size=5mm,font=\footnotesize,rectangle,rounded corners=2.5mm,inner sep=0.2mm]
\tikzstyle{gray rect ddot}=[draw=black,fill=gray!40!white,doubled,minimum size=6mm,font=\footnotesize,rectangle,rounded corners=3mm]

\tikzstyle{gray phase dot}=[dot,fill=gray!40!white,phase dimensions]
\tikzstyle{gray phase ddot}=[ddot,fill=gray!40!white,dphase dimensions]
\tikzstyle{grey phase dot}=[gray phase dot]
\tikzstyle{grey phase ddot}=[gray phase ddot]

\tikzstyle{small phase dimensions}=[minimum size=4mm,font=\tiny,rectangle,rounded corners=2mm,inner sep=0.2mm,outer sep=-2mm]
\tikzstyle{small dphase dimensions}=[minimum size=4mm,font=\tiny,rectangle,rounded corners=2mm,inner sep=0.2mm,outer sep=-2mm]

\tikzstyle{small gray phase dot}=[dot,fill=gray!40!white,small phase dimensions]
\tikzstyle{small gray phase ddot}=[ddot,fill=gray!40!white,small dphase dimensions]

\tikzstyle{small map}=[draw,shape=rectangle,minimum height=4mm,minimum width=4mm,fill=white]

\tikzstyle{cnot}=[fill=white,shape=circle,inner sep=-1.4pt]

\tikzstyle{asym hadamard}=[fill=white,draw,shape=NEbox,inner sep=0.6mm,font=\footnotesize,minimum height=4mm]
\tikzstyle{asym hadamard conj}=[fill=white,draw,shape=NWbox,inner sep=0.6mm,font=\footnotesize,minimum height=4mm]
\tikzstyle{asym hadamard dag}=[fill=white,draw,shape=SEbox,inner sep=0.6mm,font=\footnotesize,minimum height=4mm]

\tikzstyle{hadamard}=[fill=white,draw,inner sep=0.6mm,font=\footnotesize,minimum height=4mm,minimum width=4mm]
\tikzstyle{small hadamard}=[fill=white,draw,inner sep=0.6mm,minimum height=1.5mm,minimum width=1.5mm]
\tikzstyle{dhadamard}=[hadamard,doubled]
\tikzstyle{small dhadamard}=[small hadamard,doubled]
\tikzstyle{small dhadamard rotate}=[small hadamard,doubled,rotate=45]
\tikzstyle{antipode}=[white dot,inner sep=0.3mm,font=\footnotesize]

\tikzstyle{scalar}=[diamond,draw,inner sep=0.5pt,font=\small]
\tikzstyle{dscalar}=[diamond,doubled, draw,inner sep=0.5pt,font=\small]

\tikzstyle{small box}=[rectangle,inline text,fill=white,draw,minimum height=5mm,yshift=-0.5mm,minimum width=5mm,font=\small]
\tikzstyle{small gray box}=[small box,fill=gray!30]
\tikzstyle{medium box}=[rectangle,inline text,fill=white,draw,minimum height=5mm,yshift=-0.5mm,minimum width=10mm,font=\small]
\tikzstyle{square box}=[small box] 
\tikzstyle{medium gray box}=[small box,fill=gray!30]
\tikzstyle{semilarge box}=[rectangle,inline text,fill=white,draw,minimum height=5mm,yshift=-0.5mm,minimum width=12.5mm,font=\small]
\tikzstyle{large box}=[rectangle,inline text,fill=white,draw,minimum height=5mm,yshift=-0.5mm,minimum width=15mm,font=\small]
\tikzstyle{large gray box}=[small box,fill=gray!30]

\tikzstyle{Bayes box}=[rectangle,fill=black,draw, minimum height=3mm, minimum width=3mm]

\tikzstyle{gray square point}=[small box,fill=gray!50]

\tikzstyle{dphase box white}=[dhadamard]
\tikzstyle{dphase box gray}=[dhadamard,fill=gray!50!white]

\tikzstyle{point}=[regular polygon,regular polygon sides=3,draw,scale=0.75,inner sep=-0.5pt,minimum width=9mm,fill=white,regular polygon rotate=180]
\tikzstyle{copoint}=[regular polygon,regular polygon sides=3,draw,scale=0.75,inner sep=-0.5pt,minimum width=9mm,fill=white]
\tikzstyle{dpoint}=[point,doubled]
\tikzstyle{dcopoint}=[copoint,doubled]

\tikzstyle{wide copoint}=[fill=white,draw,shape=isosceles triangle,shape border rotate=90,isosceles triangle stretches=true,inner sep=0pt,minimum width=1.5cm,minimum height=6.12mm]
\tikzstyle{wide point}=[fill=white,draw,shape=isosceles triangle,shape border rotate=-90,isosceles triangle stretches=true,inner sep=0pt,minimum width=1.5cm,minimum height=6.12mm,yshift=-0.0mm]
\tikzstyle{wide point plus}=[fill=white,draw,shape=isosceles triangle,shape border rotate=-90,isosceles triangle stretches=true,inner sep=0pt,minimum width=1.74cm,minimum height=7mm,yshift=-0.0mm]

\tikzstyle{wide dpoint}=[fill=white,doubled,draw,shape=isosceles triangle,shape border rotate=-90,isosceles triangle stretches=true,inner sep=0pt,minimum width=1.5cm,minimum height=6.12mm,yshift=-0.0mm]
\tikzstyle{wide dcopoint}=[fill=white,doubled,draw,shape=isosceles triangle,shape border rotate=90,isosceles triangle stretches=true,inner sep=0pt,minimum width=1.5cm,minimum height=6.12mm,yshift=-0.0mm]

\tikzstyle{tinypoint}=[regular polygon,regular polygon sides=3,draw,scale=0.55,inner sep=-0.15pt,minimum width=6mm,fill=white,regular polygon rotate=180]

\tikzstyle{white point}=[point]
\tikzstyle{white dpoint}=[dpoint]
\tikzstyle{green point}=[white point] 
\tikzstyle{white copoint}=[copoint]
\tikzstyle{gray point}=[point,fill=gray!40!white]
\tikzstyle{gray dpoint}=[gray point,doubled]
\tikzstyle{red point}=[gray point] 
\tikzstyle{gray copoint}=[copoint,fill=gray!40!white]
\tikzstyle{gray dcopoint}=[gray copoint,doubled]

\tikzstyle{white point guide}=[regular polygon,regular polygon sides=3,font=\scriptsize,draw,scale=0.65,inner sep=-0.5pt,minimum width=9mm,fill=white,regular polygon rotate=180]

\tikzstyle{black point}=[point,fill=black,font=\color{white}]
\tikzstyle{black copoint}=[copoint,fill=black,font=\color{white}]

\tikzstyle{tiny gray point}=[tinypoint,fill=gray!40!white]

\tikzstyle{diredge}=[->]
\tikzstyle{ddiredge}=[<->]
\tikzstyle{rdiredge}=[<-]
\tikzstyle{thickdiredge}=[->, very thick]
\tikzstyle{pointer edge}=[->,very thick,gray]
\tikzstyle{pointer edge part}=[very thick,gray]
\tikzstyle{dashed edge}=[dashed]
\tikzstyle{thick dashed edge}=[very thick,dashed]
\tikzstyle{thick gray dashed edge}=[thick dashed edge,gray!40]
\tikzstyle{thick map edge}=[very thick,|->]

\makeatletter
\newcommand{\boxshape}[3]{%
\pgfdeclareshape{#1}{
\inheritsavedanchors[from=rectangle] 
\inheritanchorborder[from=rectangle]
\inheritanchor[from=rectangle]{center}
\inheritanchor[from=rectangle]{north}
\inheritanchor[from=rectangle]{south}
\inheritanchor[from=rectangle]{west}
\inheritanchor[from=rectangle]{east}
\backgroundpath{
\southwest \pgf@xa=\pgf@x \pgf@ya=\pgf@y
\northeast \pgf@xb=\pgf@x \pgf@yb=\pgf@y

\@tempdima=#2
\@tempdimb=#3

\pgfpathmoveto{\pgfpoint{\pgf@xa - 5pt + \@tempdima}{\pgf@ya}}
\pgfpathlineto{\pgfpoint{\pgf@xa - 5pt - \@tempdima}{\pgf@yb}}
\pgfpathlineto{\pgfpoint{\pgf@xb + 5pt + \@tempdimb}{\pgf@yb}}
\pgfpathlineto{\pgfpoint{\pgf@xb + 5pt - \@tempdimb}{\pgf@ya}}
\pgfpathlineto{\pgfpoint{\pgf@xa - 5pt + \@tempdima}{\pgf@ya}}
\pgfpathclose
}
}}

\boxshape{NEbox}{0pt}{5pt}
\boxshape{SEbox}{0pt}{-5pt}
\boxshape{NWbox}{5pt}{0pt}
\boxshape{SWbox}{-5pt}{0pt}
\boxshape{EBox}{-3pt}{3pt}
\boxshape{WBox}{3pt}{-3pt}
\makeatother

\tikzstyle{cloud}=[shape=cloud,draw,minimum width=1.5cm,minimum height=1.5cm]

\tikzstyle{map}=[draw,shape=NEbox,inner sep=2pt,minimum height=6mm,fill=white]
\tikzstyle{dashedmap}=[draw,dashed,shape=NEbox,inner sep=2pt,minimum height=6mm,fill=white]
\tikzstyle{mapdag}=[draw,shape=SEbox,inner sep=2pt,minimum height=6mm,fill=white]
\tikzstyle{mapadj}=[draw,shape=SEbox,inner sep=2pt,minimum height=6mm,fill=white]
\tikzstyle{maptrans}=[draw,shape=SWbox,inner sep=2pt,minimum height=6mm,fill=white]
\tikzstyle{mapconj}=[draw,shape=NWbox,inner sep=2pt,minimum height=6mm,fill=white]

\tikzstyle{medium map}=[draw,shape=NEbox,inner sep=2pt,minimum height=6mm,fill=white,minimum width=7mm]
\tikzstyle{medium map dag}=[draw,shape=SEbox,inner sep=2pt,minimum height=6mm,fill=white,minimum width=7mm]
\tikzstyle{medium map adj}=[draw,shape=SEbox,inner sep=2pt,minimum height=6mm,fill=white,minimum width=7mm]
\tikzstyle{medium map trans}=[draw,shape=SWbox,inner sep=2pt,minimum height=6mm,fill=white,minimum width=7mm]
\tikzstyle{medium map conj}=[draw,shape=NWbox,inner sep=2pt,minimum height=6mm,fill=white,minimum width=7mm]
\tikzstyle{semilarge map}=[draw,shape=NEbox,inner sep=2pt,minimum height=6mm,fill=white,minimum width=9.5mm]
\tikzstyle{semilarge map trans}=[draw,shape=SWbox,inner sep=2pt,minimum height=6mm,fill=white,minimum width=9.5mm]
\tikzstyle{semilarge map adj}=[draw,shape=SEbox,inner sep=2pt,minimum height=6mm,fill=white,minimum width=9.5mm]
\tikzstyle{semilarge map dag}=[draw,shape=SEbox,inner sep=2pt,minimum height=6mm,fill=white,minimum width=9.5mm]
\tikzstyle{semilarge map conj}=[draw,shape=NWbox,inner sep=2pt,minimum height=6mm,fill=white,minimum width=9.5mm]
\tikzstyle{large map}=[draw,shape=NEbox,inner sep=2pt,minimum height=6mm,fill=white,minimum width=12mm]
\tikzstyle{large map conj}=[draw,shape=NWbox,inner sep=2pt,minimum height=6mm,fill=white,minimum width=12mm]
\tikzstyle{very large map}=[draw,shape=NEbox,inner sep=2pt,minimum height=6mm,fill=white,minimum width=17mm]

\tikzstyle{medium dmap}=[draw,doubled,shape=NEbox,inner sep=2pt,minimum height=6mm,fill=white,minimum width=7mm]
\tikzstyle{medium dmap dag}=[draw,doubled,shape=SEbox,inner sep=2pt,minimum height=6mm,fill=white,minimum width=7mm]
\tikzstyle{medium dmap adj}=[draw,doubled,shape=SEbox,inner sep=2pt,minimum height=6mm,fill=white,minimum width=7mm]
\tikzstyle{medium dmap trans}=[draw,doubled,shape=SWbox,inner sep=2pt,minimum height=6mm,fill=white,minimum width=7mm]
\tikzstyle{medium dmap conj}=[draw,doubled,shape=NWbox,inner sep=2pt,minimum height=6mm,fill=white,minimum width=7mm]
\tikzstyle{semilarge dmap}=[draw,doubled,shape=NEbox,inner sep=2pt,minimum height=6mm,fill=white,minimum width=9.5mm]
\tikzstyle{semilarge dmap trans}=[draw,doubled,shape=SWbox,inner sep=2pt,minimum height=6mm,fill=white,minimum width=9.5mm]
\tikzstyle{semilarge dmap adj}=[draw,doubled,shape=SEbox,inner sep=2pt,minimum height=6mm,fill=white,minimum width=9.5mm]
\tikzstyle{semilarge dmap dag}=[draw,doubled,shape=SEbox,inner sep=2pt,minimum height=6mm,fill=white,minimum width=9.5mm]
\tikzstyle{semilarge dmap conj}=[draw,doubled,shape=NWbox,inner sep=2pt,minimum height=6mm,fill=white,minimum width=9.5mm]
\tikzstyle{large dmap}=[draw,doubled,shape=NEbox,inner sep=2pt,minimum height=6mm,fill=white,minimum width=12mm]
\tikzstyle{large dmap conj}=[draw,doubled,shape=NWbox,inner sep=2pt,minimum height=6mm,fill=white,minimum width=12mm]
\tikzstyle{large dmap trans}=[draw,doubled,shape=SWbox,inner sep=2pt,minimum height=6mm,fill=white,minimum width=12mm]
\tikzstyle{large dmap adj}=[draw,doubled,shape=SEbox,inner sep=2pt,minimum height=6mm,fill=white,minimum width=12mm]
\tikzstyle{large dmap dag}=[draw,doubled,shape=SEbox,inner sep=2pt,minimum height=6mm,fill=white,minimum width=12mm]
\tikzstyle{very large dmap}=[draw,doubled,shape=NEbox,inner sep=2pt,minimum height=6mm,fill=white,minimum width=19.5mm]

\tikzstyle{muxbox}=[draw,shape=rectangle,minimum height=3mm,minimum width=3mm,fill=white]
\tikzstyle{dmuxbox}=[muxbox,doubled]

\tikzstyle{box}=[draw,shape=rectangle,inner sep=2pt,minimum height=6mm,minimum width=6mm,fill=white]
\tikzstyle{dbox}=[draw,doubled,shape=rectangle,inner sep=2pt,minimum height=6mm,minimum width=6mm,fill=white]
\tikzstyle{dmap}=[draw,doubled,shape=NEbox,inner sep=2pt,minimum height=6mm,fill=white]
\tikzstyle{dmapdag}=[draw,doubled,shape=SEbox,inner sep=2pt,minimum height=6mm,fill=white]
\tikzstyle{dmapadj}=[draw,doubled,shape=SEbox,inner sep=2pt,minimum height=6mm,fill=white]
\tikzstyle{dmaptrans}=[draw,doubled,shape=SWbox,inner sep=2pt,minimum height=6mm,fill=white]
\tikzstyle{dmapconj}=[draw,doubled,shape=NWbox,inner sep=2pt,minimum height=6mm,fill=white]

\tikzstyle{ddmap}=[draw,doubled,dashed,shape=NEbox,inner sep=2pt,minimum height=6mm,fill=white]
\tikzstyle{ddmapdag}=[draw,doubled,dashed,shape=SEbox,inner sep=2pt,minimum height=6mm,fill=white]
\tikzstyle{ddmapadj}=[draw,doubled,dashed,shape=SEbox,inner sep=2pt,minimum height=6mm,fill=white]
\tikzstyle{ddmaptrans}=[draw,doubled,dashed,shape=SWbox,inner sep=2pt,minimum height=6mm,fill=white]
\tikzstyle{ddmapconj}=[draw,doubled,dashed,shape=NWbox,inner sep=2pt,minimum height=6mm,fill=white]

\boxshape{sNEbox}{0pt}{3pt}
\boxshape{sSEbox}{0pt}{-3pt}
\boxshape{sNWbox}{3pt}{0pt}
\boxshape{sSWbox}{-3pt}{0pt}
\tikzstyle{smap}=[draw,shape=sNEbox,fill=white]
\tikzstyle{smapdag}=[draw,shape=sSEbox,fill=white]
\tikzstyle{smapadj}=[draw,shape=sSEbox,fill=white]
\tikzstyle{smaptrans}=[draw,shape=sSWbox,fill=white]
\tikzstyle{smapconj}=[draw,shape=sNWbox,fill=white]

\tikzstyle{dsmap}=[draw,dashed,shape=sNEbox,fill=white]
\tikzstyle{dsmapdag}=[draw,dashed,shape=sSEbox,fill=white]
\tikzstyle{dsmaptrans}=[draw,dashed,shape=sSWbox,fill=white]
\tikzstyle{dsmapconj}=[draw,dashed,shape=sNWbox,fill=white]

\boxshape{mNEbox}{0pt}{10pt}
\boxshape{mSEbox}{0pt}{-10pt}
\boxshape{mNWbox}{10pt}{0pt}
\boxshape{mSWbox}{-10pt}{0pt}
\tikzstyle{mmap}=[draw,shape=mNEbox]
\tikzstyle{mmapdag}=[draw,shape=mSEbox]
\tikzstyle{mmaptrans}=[draw,shape=mSWbox]
\tikzstyle{mmapconj}=[draw,shape=mNWbox]

\tikzstyle{mmapgray}=[draw,fill=gray!40!white,shape=mNEbox]
\tikzstyle{smapgray}=[draw,fill=gray!40!white,shape=sNEbox]

\makeatletter
\pgfdeclareshape{cornerpoint}{
\inheritsavedanchors[from=rectangle] 
\inheritanchorborder[from=rectangle]
\inheritanchor[from=rectangle]{center}
\inheritanchor[from=rectangle]{north}
\inheritanchor[from=rectangle]{south}
\inheritanchor[from=rectangle]{west}
\inheritanchor[from=rectangle]{east}
\backgroundpath{
\southwest \pgf@xa=\pgf@x \pgf@ya=\pgf@y
\northeast \pgf@xb=\pgf@x \pgf@yb=\pgf@y

\pgfmathsetmacro{\pgf@shorten@left}{\pgfkeysvalueof{/tikz/shorten left}}
\pgfmathsetmacro{\pgf@shorten@right}{\pgfkeysvalueof{/tikz/shorten right}}

\pgfpathmoveto{\pgfpoint{0.5 * (\pgf@xa + \pgf@xb)}{\pgf@ya - 5pt}}
\pgfpathlineto{\pgfpoint{\pgf@xa - 8pt + \pgf@shorten@left}{\pgf@yb - 1.5 * \pgf@shorten@left}}
\pgfpathlineto{\pgfpoint{\pgf@xa - 8pt + \pgf@shorten@left}{\pgf@yb}}
\pgfpathlineto{\pgfpoint{\pgf@xb + 8pt - \pgf@shorten@right}{\pgf@yb}}
\pgfpathlineto{\pgfpoint{\pgf@xb + 8pt - \pgf@shorten@right}{\pgf@yb - 1.5 * \pgf@shorten@right}}
\pgfpathclose
}
}

\pgfdeclareshape{cornercopoint}{
\inheritsavedanchors[from=rectangle] 
\inheritanchorborder[from=rectangle]
\inheritanchor[from=rectangle]{center}
\inheritanchor[from=rectangle]{north}
\inheritanchor[from=rectangle]{south}
\inheritanchor[from=rectangle]{west}
\inheritanchor[from=rectangle]{east}
\backgroundpath{
\southwest \pgf@xa=\pgf@x \pgf@ya=\pgf@y
\northeast \pgf@xb=\pgf@x \pgf@yb=\pgf@y

\pgfmathsetmacro{\pgf@shorten@left}{\pgfkeysvalueof{/tikz/shorten left}}
\pgfmathsetmacro{\pgf@shorten@right}{\pgfkeysvalueof{/tikz/shorten right}}

\pgfpathmoveto{\pgfpoint{0.5 * (\pgf@xa + \pgf@xb)}{\pgf@yb + 5pt}}
\pgfpathlineto{\pgfpoint{\pgf@xa - 8pt + \pgf@shorten@left}{\pgf@ya + 1.5 * \pgf@shorten@left}}
\pgfpathlineto{\pgfpoint{\pgf@xa - 8pt + \pgf@shorten@left}{\pgf@ya}}
\pgfpathlineto{\pgfpoint{\pgf@xb + 8pt - \pgf@shorten@right}{\pgf@ya}}
\pgfpathlineto{\pgfpoint{\pgf@xb + 8pt - \pgf@shorten@right}{\pgf@ya + 1.5 * \pgf@shorten@right}}
\pgfpathclose
}
}

\makeatother

\pgfkeyssetvalue{/tikz/shorten left}{0pt}
\pgfkeyssetvalue{/tikz/shorten right}{0pt}

\tikzstyle{kpoint common}=[draw,fill=white,inner sep=1pt,minimum height=4mm]
\tikzstyle{kpoint}=[shape=cornerpoint,shorten left=5pt,kpoint common]
\tikzstyle{kpoint adjoint}=[shape=cornercopoint,shorten left=5pt,kpoint common]
\tikzstyle{kpoint conjugate}=[shape=cornerpoint,shorten right=5pt,kpoint common]
\tikzstyle{kpoint transpose}=[shape=cornercopoint,shorten right=5pt,kpoint common]
\tikzstyle{kpoint symm}=[shape=cornerpoint,shorten left=5pt,shorten right=5pt,kpoint common]

\tikzstyle{black kpoint}=[shape=cornerpoint,shorten left=5pt,kpoint common,fill=black,font=\color{white}]
\tikzstyle{black kpoint adjoint}=[shape=cornercopoint,shorten left=5pt,kpoint common,fill=black,font=\color{white}]
\tikzstyle{black kpointadj}=[shape=cornercopoint,shorten left=5pt,kpoint common,fill=black,font=\color{white}]

\tikzstyle{black dkpoint}=[shape=cornerpoint,shorten left=5pt,kpoint common,fill=black, doubled,font=\color{white}]
\tikzstyle{black dkpoint adjoint}=[shape=cornercopoint,shorten left=5pt,kpoint common,fill=black, doubled,font=\color{white}]
\tikzstyle{black dkpointadj}=[shape=cornercopoint,shorten left=5pt,kpoint common,fill=black, doubled,font=\color{white}]

\tikzstyle{kpointdag}=[kpoint adjoint]
\tikzstyle{kpointadj}=[kpoint adjoint]
\tikzstyle{kpointconj}=[kpoint conjugate]
\tikzstyle{kpointtrans}=[kpoint transpose]

\tikzstyle{big kpoint}=[kpoint, minimum width=1.2 cm, minimum height=8mm, inner sep=4pt, text depth=3mm]

\tikzstyle{wide kpoint}=[kpoint, minimum width=1 cm, inner sep=2pt]
\tikzstyle{wide kpointdag}=[kpointdag, minimum width=1 cm, inner sep=2pt]
\tikzstyle{wide kpointconj}=[kpointconj, minimum width=1 cm, inner sep=2pt]
\tikzstyle{wide kpointtrans}=[kpointtrans, minimum width=1 cm, inner sep=2pt]

\tikzstyle{gray kpoint}=[kpoint,fill=gray!50!white]
\tikzstyle{gray kpointdag}=[kpointdag,fill=gray!50!white]
\tikzstyle{gray kpointadj}=[kpointadj,fill=gray!50!white]
\tikzstyle{gray kpointconj}=[kpointconj,fill=gray!50!white]
\tikzstyle{gray kpointtrans}=[kpointtrans,fill=gray!50!white]

\tikzstyle{gray dkpoint}=[kpoint,fill=gray!50!white,doubled]
\tikzstyle{gray dkpointdag}=[kpointdag,fill=gray!50!white,doubled]
\tikzstyle{gray dkpointadj}=[kpointadj,fill=gray!50!white,doubled]
\tikzstyle{gray dkpointconj}=[kpointconj,fill=gray!50!white,doubled]
\tikzstyle{gray dkpointtrans}=[kpointtrans,fill=gray!50!white,doubled]

\tikzstyle{white label}=[draw,fill=white,rectangle,inner sep=0.7 mm]
\tikzstyle{gray label}=[draw,fill=gray!50!white,rectangle,inner sep=0.7 mm]
\tikzstyle{black label}=[draw,fill=black,rectangle,inner sep=0.7 mm]

\tikzstyle{dkpoint}=[kpoint,doubled]
\tikzstyle{wide dkpoint}=[wide kpoint,doubled]
\tikzstyle{dkpointdag}=[kpoint adjoint,doubled]
\tikzstyle{wide dkpointdag}=[wide kpointdag,doubled]
\tikzstyle{dkcopoint}=[kpoint adjoint,doubled]
\tikzstyle{dkpointadj}=[kpoint adjoint,doubled]
\tikzstyle{dkpointconj}=[kpoint conjugate,doubled]
\tikzstyle{dkpointtrans}=[kpoint transpose,doubled]

\tikzstyle{kscalar}=[kpoint common, shape=EBox, inner xsep=-1pt, inner ysep=3pt,font=\small]
\tikzstyle{kscalarconj}=[kpoint common, shape=WBox, inner xsep=-1pt, inner ysep=3pt,font=\small]

 \tikzstyle{upground}=[circuit ee IEC,thick,ground,rotate=90,scale=2.5]
 \tikzstyle{downground}=[circuit ee IEC,thick,ground,rotate=-90,scale=2.5]
 \tikzstyle{bigground}=[regular polygon,regular polygon sides=3,draw=gray,scale=0.50,inner sep=-0.5pt,minimum width=10mm,fill=gray]

\tikzstyle{arrs}=[-latex,font=\small,auto]
\tikzstyle{arrow plain}=[arrs]
\tikzstyle{arrow dashed}=[dashed,arrs]
\tikzstyle{arrow bold}=[very thick,arrs]
\tikzstyle{arrow hide}=[draw=white!0,-]
\tikzstyle{arrow reverse}=[latex-]
\tikzstyle{cdnode}=[]

\definecolor{hexcolor0xa9a9a9}{rgb}{0.663,0.663,0.663}
\tikzstyle{GrayLine}=[dashed,draw=hexcolor0xa9a9a9]
\tikzstyle{gray}=[dashed,draw=hexcolor0xa9a9a9]

\theoremstyle{definition}
\newtheorem{theorem}{Theorem}[section]
\newtheorem*{theorem*}{Theorem}

\newtheorem{defn}[theorem]{Definition}

\newtheorem{example}[theorem]{Example}

\newtheorem{example*}[theorem]{Example*}
\newtheorem{examples*}[theorem]{Examples*}

\newtheorem{remark*}[theorem]{Remark*}

\newcommand{\vect}[1]{\overrightarrow{#1}}
\newcommand{\lang}[1]{\textit{#1}}
\newcommand{\ConvexRel}{\mathbf{ConvexRel}}

\def\bR{\begin{color}{red}}
\def\bB{\begin{color}{blue}}
\def\bM{\begin{color}{magenta}}
\def\bC{\begin{color}{cyan}}
\def\bW{\begin{color}{white}}
\def\bBl{\begin{color}{black}}
\def\bG{\begin{color}{green}}
\def\bY{\begin{color}{yellow}}
\def\e{\end{color}\xspace}

\newcommand{\thetitle}{Internal Wiring of Cartesian Verbs and Prepositions}
\title{\thetitle}
\newcommand{\titlerunning}{\thetitle}
\newcommand{\authorrunning}{B. Coecke, M. Lewis \& D. Marsden}

\author{Bob Coecke\institute{University of Oxford}\email{bob.coecke@cs.ox.ac.uk} \and Martha Lewis \institute{ILLC, University of Amsterdam}\email{m.a.f.lewis@uva.nl} \and Dan Marsden \institute{University of Oxford}\email{daniel.marsden@cs.ox.ac.uk}}

\begin{document}
\maketitle 

\begin{abstract}
Categorical  compositional distributional semantics (CCDS) allows one to compute the meaning of phrases and sentences from the meaning of their constituent words. A type-structure  carried over from the traditional categorial model of grammar a la Lambek becomes a `wire-structure' that mediates the interaction of word meanings.  However, CCDS has a much richer logical structure than plain categorical semantics in that certain words can also be given an `internal wiring' that  either provides their entire meaning or reduces the size their meaning space. Previous examples of internal wiring include relative pronouns and intersective adjectives.  Here we establish the same for a large class of well-behaved transitive verbs to which we refer as Cartesian verbs, and reduce the meaning space from a ternary tensor to a unary one. Some experimental evidence is also provided.
\end{abstract} 

\section{Introduction} 
How does grammar allow us to combine word meanings to form sentence meanings? An answer is provided by categorical compositional distributional semantics (CCDS) \cite{CSC, LambekvsLambek}. In CCDS, a type-structure  from traditional categorial grammar a la Lambek, in any of its various forms~\cite{Lambek0, grishin1983generalization, Lambek1}, is carried over functorially to a corresponding category of meaning spaces. The image of this grammatical reduction becomes a morphism mapping meanings of words into meanings of sentences.  Factoring out the symbolic baggage of category theory, one obtains a `wire-structure'~\cite{CatsII, CKpaperI, CKbook} that mediates the interaction of word meanings. Interestingly, CCDS was initially inspired by a diagrammatic formulations of quantum theory~\cite{teleling, DBLP:books/daglib/p/Coecke17}, and yet for a number of tasks it outperforms traditional models of language, see for example the contrast between~\cite{mitchelllapata} and~\cite{KartSadr, GrefSadr}.  

A further benefit of CCDS is that it is model-independent, in the sense that any category of meaning spaces which is endowed with an appropriate structure will do.  Examples of practical interest include vector spaces~\cite{CSC}, relations~\cite{CSC}, density matrices~\cite{calco2015, bankova2016graded}, and conceptual spaces~\cite{ConcSpacI} a la G\"ardenfors~\cite{gardenfors}.

Another feature of CCDS is that one can also assign `internal wiring' to special functional words that either capture their entire meaning, or identify substructure within more general classes of words.  Examples of the former are the words ``and''~\cite{DBLP:journals/corr/Kartsaklis16}, ``does'', ``not''~\cite{CSC}, and relative pronouns~\cite{FrobMeanI, FrobMeanII} which go beyond wires and involve `multi-wires' or `spiders'~\cite{CQMII, CKbook}. Intersective adjectives are examples of a class of words with internal structure~\cite{ConcSpacI}.  

One practical downside of CCDS is that words now have types which sometimes can be quite large, resulting in humongous sizes of meaning spaces.  For example, transitive verbs have a composite type consisting of three basic types, so if we talk vector spaces and each basic type has dimension 1,000, the verb lives in a space of dimension 1,000,000,000!  This can be reduced by means of an internal wiring that decomposes the big vector into smaller ones, and that's exactly what we do in this paper.  We define Cartesian verbs as certain well-behaving verbs that admit a specific internal wiring. As a corollary, we derive an internal wiring for prepositions.  We examine some interesting interaction with relative pronouns, and provide some empirical evidence of the validity of our decomposition.

Due to lack of space we will assume that the reader has some familiarity with diagrammatic notation like the one employed in CCDS.  Details can be found in~\cite{CKpaperI, CQMII, CKbook}.

\section{Background}
 
\subsection{Grammatical wirings}
The mathematical formulation of grammar~\cite{Ajdukiewicz, Bar-Hillel, Lambek0} teaches us that the fundamental entities making up phrases and sentences are not  words, but some basic grammatical types. Both the noun-type $n$, and the type of whole sentences~$s$ are examples of these basic types. A transitive verb is not a basic type, it is a composite made up of two noun-types and one sentence-type.

In this paper, our grammars will follow~\cite{Lambek1}, yielding a elegant diagrammatic formulation. For each basic type~$x$ there are corresponding basic `anti-types', and there are in fact two of these, which we denote~${}^{-1}x$ and~$x^{-1}$.  Then, in English, a transitive verb has type ${}^{-1}n \cdot s \cdot n^{-1}$. To understand this type, consider a transitive verb like `hate'. Simply saying `hate' doesn't convey any useful information, until we also specify `whom' hates `whom'.  That's exactly the role of the anti-types: they specify that in order to form a meaningful sentence they need to be ``cancelled out'', in this case by composing with a noun on both the left and right: 
\[
\underbrace{Alice}_{n} \underbrace{hates}_{{}^{-1}n \cdot s \cdot n^{-1}}  \underbrace{Bob}_n
\]  
Both $n \cdot {}^{-1}n$ and $n^{-1} \cdot n$ cancel, and what remains is $s$, confirming that ``Alice hates Bob'' is a grammatically well-typed sentence.  All of this is governed by an algebraic structure called a \em pregroup\em.

Instead of algebraically, one can also depict the cancelations using wires, as follows:    
\ctikzfig{hates}
For a more complex sentence like ``Alice does not like Bob" also this wiring will be more complex \cite{CSC}:
\ctikzfig{hatescomplgram}
but the idea remains the same.

The main idea of CCDS is to think of these wires not just as a representation of the computation that establishes that we are dealing with a grammatically well-formed sentence, but as a representation of how the meanings of the words making up the sentence interact.  Abstractly representing word-meanings as follows:
\ctikzfig{hates2}   
where the triangles represent the meanings, and the outgoing wires correspond to the type.  We can now apply the wiring to these as follows: 
\ctikzfig{hates3}   
The wires now `feed' the object and the subject into the verb in order to produce the meaning of the sentence as a whole. 

Concretely, following the most common current paradigm in Natural Language Processing (NLP) \cite{Schuetze}, one may think of these meanings as vectors in a vector space and then the wiring correspond to a certain linear map mostly made up of inner-products \cite{CSC}.  Applying this linear maps to the tensor product of the meaning vectors then yields another vector which we conceive as the meaning of the sentence \cite{CSC}.  However, one could consider many other models of meaning as long as they carry same abstract mathematical structure as the grammar, in this case that of a compact closed category. Useful examples are density matrices~\cite{bankova2016graded} and subsets of certain convex spaces~\cite{ConcSpacI}, but a plethora of other models are available \cite{DBLP:conf/wollic/CoeckeGLM17}. 

\subsection{Internal wirings of meanings}  
\label{sec:wirings}
Not only grammatical structure, but also certain meanings themselves can be represented using wiring. An example of this~\cite{PrelSadr, CSC} is the sentence already mentioned above:
\ctikzfig{hatescompl}
We see here that both ``does" and ``not" have been represented involving wires:
\ctikzfig{hatescomplclever}
with ``does" being entirely made up of wires and ``not" also involving a $\neg$-labeled box which represents negation of the meaning, something which obviously will not exist in every possible model of meaning.  That these wirings are the right ones becomes clear when we yank the wires in order to obtain:
\ctikzfig{hatescompl2}
where we also transposed ``like'' in order to turn it into a function-like  box. 
So we obtain the negation of ``like" applied to the pair (``Alice",``Bob"), which is indeed the intended meaning of the sentence.
  
Another somewhat more sophisticated example taken from \cite{FrobMeanI, FrobMeanII} involves the representation of relative pronouns by `multi-wires' or `spiders'  \cite{CQMII, CKbook}:
\ctikzfig{relpron}
The spider with three legs should be thought of as a conjunction, while the one with a single leg kills the sentence type.  Simplification now yields:  
\ctikzfig{relpron2} 
So we obtain a conjunction of ``she" (i.e.~`being female') and ``... hates Bob'', which is indeed again the intended meaning of the sentence. 
 
These multi-wires can also be used to encode certain kinds of well-behaved adjectives called \em intersective adjectives \em \cite{KampPartee1995}.  These are adjectives which leave a noun unaltered except for specifying an additional property, e.g.~``red car'', ``hot food'' or ``sad Bob", as opposed to ``crashed car'',  ``rotten food'' or ``dead Bob". While a general adjective has a composite type e.g.:
\ctikzfig{adj1}  
in the case of intersective adjective its meaning can be reduced to a single wire \cite{ConcSpacI}:  
\ctikzfig{adj2}
which indeed yields a conjunction:
\ctikzfig{adj3}

\subsection{Related work}
The model we present here has similarities to models presented in \cite{KartsaklisSadrzadeh2014} and can perhaps be seen as a generalization of those models. In that work, the authors notice an intriguing aspect of verb and adjective tensors learnt from corpus data. This is that they can be approximated by separable tensors, which considerably simplifies the the interactions between words. In the worst case, this separability means that the meanings of phrases  are dependent only on the meaning of the functional word:
\[
\begin{gathered}\begin{tikzpicture}
	\begin{pgfonlayer}{nodelayer}
		\node [style=none] (0) at (-3, 0) {};
		\node [style=none] (1) at (3, 0) {};
		\node [style=none] (2) at (0, 0) {};
		\node [style=none] (3) at (1, 0) {};
		\node [style=none] (4) at (1, -0.5) {};
		\node [style=kpoint] (5) at (-3, -0.75) {\footnotesize$A$};
		\node [style=kpoint] (6) at (3, -0.75) {\footnotesize$B$};
		\node [style=kpoint, minimum width=1.3225 cm, inner sep=2 pt] (7) at (0, -0.75) {\footnotesize$hates$};
		\node [style=none] (8) at (-1, 0) {};
		\node [style=none] (9) at (-1, -0.5) {};
		\node [style=none] (10) at (0, 1.5) {};  
		\node [style=none] (11) at (1, 0) {};
		\node [style=none] (12) at (-3, 0) {};
		\node [style=none] (13) at (-1, 0) {};
		\node [style=none] (14) at (-1, 0) {};  
		\node [style=none] (15) at (0, 0) {};
		\node [style=none] (16) at (0, 0) {};
		\node [style=none] (17) at (1, 0) {};
		\node [style=none] (18) at (3, 0) {};
	\end{pgfonlayer}
	\begin{pgfonlayer}{edgelayer}
		\draw [style=swap] (3.center) to (4.center);
		\draw [style=swap] (1.center) to (6);
		\draw [style=swap] (0.center) to (5);
		\draw [style=swap] (2.center) to (7);
		\draw [style=swap] (8.center) to (9.center);
		\draw [style=swap] (10.center) to (15.center);
		\draw [style=swap, bend left=90, looseness=1.75] (12.center) to (13.center);
		\draw [style=swap, bend left=90, looseness=1.75] (11.center) to (18.center);
	\end{pgfonlayer}
\end{tikzpicture}\end{gathered} \quad
= \quad \begin{gathered}\begin{tikzpicture}
	\begin{pgfonlayer}{nodelayer}
		\node [style=none] (0) at (-4, 0) {};
		\node [style=none] (1) at (4, 0) {};
		\node [style=none] (2) at (0, 0) {};
		\node [style=none] (3) at (2, 0) {};
		\node [style=none] (4) at (2, -0.75) {};
		\node [style=kpoint] (5) at (-4, -0.75) {\footnotesize$A$};
		\node [style=kpoint] (6) at (4, -0.75) {\footnotesize$B$};
		\node [style=none] (7) at (-2, 0) {};
		\node [style=none] (8) at (-2, -0.75) {};
		\node [style=none] (9) at (0, 1.5) {};
		\node [style=none] (10) at (2, 0) {};
		\node [style=none] (11) at (-4, 0) {};
		\node [style=none] (12) at (-2, 0) {};
		\node [style=none] (13) at (-2, 0) {};
		\node [style=none] (14) at (0, 0) {};
		\node [style=none] (15) at (0, 0) {};
		\node [style=none] (16) at (2, 0) {};
		\node [style=none] (17) at (4, 0) {};
		\node [style=kpoint] (18) at (0, -0.75) {\footnotesize$h_m$};
		\node [style=kpoint] (19) at (2, -0.75) {\footnotesize$h_r$};
		\node [style=kpoint] (20) at (-2, -0.75) {\footnotesize$h_l$};
	\end{pgfonlayer}
	\begin{pgfonlayer}{edgelayer}
		\draw [style=swap] (3.center) to (4.center);
		\draw [style=swap] (1.center) to (6);
		\draw [style=swap] (0.center) to (5);
		\draw [style=swap] (7.center) to (8.center);
		\draw [style=swap] (9.center) to (14.center);
		\draw [style=swap, bend left=90, looseness=1.75] (11.center) to (12.center);
		\draw [style=swap, bend left=90, looseness=1.75] (10.center) to (17.center);
		\draw (15.center) to (18);
	\end{pgfonlayer}
\end{tikzpicture}\end{gathered}\quad
= \quad \begin{gathered}\begin{tikzpicture}
	\begin{pgfonlayer}{nodelayer}
		\node [style=none] (0) at (0, 0) {};
		\node [style=none] (1) at (0, 1.5) {};
		\node [style=none] (2) at (0, 0) {};
		\node [style=none] (3) at (0, 0) {};
		\node [style=kpoint] (4) at (0, -0.75) {\footnotesize$hates_m$};
	\end{pgfonlayer}
	\begin{pgfonlayer}{edgelayer}
		\draw [style=swap] (1.center) to (2.center);
		\draw (3.center) to (4);
	\end{pgfonlayer}
\end{tikzpicture}\end{gathered}
 \]
with the consequence that the information about who hates who is lost. \cite{KartsaklisSadrzadeh2014} propose a number of ways of building verb tensors which mitigate this aspect. Firstly, a verb can be built as the sum of tensor products of pairs of subject and object, first proposed in \cite{GrefSadr}. Then each dimension may be copied using the Frobenius operator. This gives a sentence space consisting of $N\otimes N$. Since the verb tensor may be approximated by a separable tensor, we obtain the following:
\[
\begin{gathered}\begin{tikzpicture}
	\begin{pgfonlayer}{nodelayer}
		\node [style=none] (0) at (-2, 1.5) {};
		\node [style=none] (1) at (-0.5, 1.5) {};
		\node [style=none] (2) at (-2, 1.5) {};
		\node [style=none] (3) at (-0.5, 1.5) {};
		\node [style=none] (4) at (-0.5, 1.5) {};
		\node [style=none] (5) at (-0.5, 1.5) {};
		\node [style=none] (6) at (-0.5, 1.5) {};
		\node [style=none] (7) at (-0.5, 1.5) {};
		\node [style=none] (8) at (-2, 1.5) {};
		\node [style=none] (9) at (-2, 1.5) {};
		\node [style=none] (10) at (-2, 1.5) {};
		\node [style=none] (11) at (-1.25, -0.5) {};
		\node [style=none] (12) at (-1.25, 0.25) {};
		\node [style=none] (13) at (-1.25, 0.25) {};
		\node [style=white dot] (14) at (-1.25, 0.25) {};
		\node [style=none] (15) at (-1.25, 0.25) {};
		\node [style=none] (16) at (1.25, 0.25) {};
		\node [style=none] (17) at (2, 1.5) {};
		\node [style=none] (18) at (1.25, -0.5) {};
		\node [style=none] (19) at (0.5, 1.5) {};
		\node [style=none] (20) at (2, 1.5) {};
		\node [style=none] (21) at (0.5, 1.5) {};
		\node [style=none] (22) at (2, 1.5) {};
		\node [style=none] (23) at (1.25, 0.25) {};
		\node [style=none] (24) at (1.25, 0.25) {};
		\node [style=none] (25) at (0.5, 1.5) {};
		\node [style=none] (26) at (2, 1.5) {};
		\node [style=none] (27) at (2, 1.5) {};
		\node [style=none] (28) at (0.5, 1.5) {};
		\node [style=white dot] (29) at (1.25, 0.25) {};
		\node [style=none] (30) at (0.5, 1.5) {};
		\node [style=none] (31) at (2, 1.5) {};
		\node [style=kpoint, minimum width=1.3225 cm, inner sep=2 pt] (32) at (0, -0.75) {\footnotesize $verb$};
	\end{pgfonlayer}
	\begin{pgfonlayer}{edgelayer}
		\draw [style=swap] (11.center) to (13.center);
		\draw [style=swap, in=165, out=-90, looseness=1.25] (10.center) to (12.center);
		\draw [style=swap, in=21, out=-90, looseness=1.25] (7.center) to (12.center);
		\draw [style=swap] (18.center) to (16.center);
		\draw [style=swap, in=165, out=-90, looseness=1.25] (21.center) to (24.center);
		\draw [style=swap, in=21, out=-90, looseness=1.25] (20.center) to (24.center);
	\end{pgfonlayer}
\end{tikzpicture}\end{gathered} \quad
\approx \quad \begin{gathered}\begin{tikzpicture}
	\begin{pgfonlayer}{nodelayer}
		\node [style=none] (0) at (-2, 1.5) {};
		\node [style=kpoint] (1) at (-1.25, -0.75) {\footnotesize $verb_l$};
		\node [style=none] (2) at (-0.5, 1.5) {};
		\node [style=none] (3) at (-2, 1.5) {};
		\node [style=none] (4) at (-0.5, 1.5) {};
		\node [style=none] (5) at (-0.5, 1.5) {};
		\node [style=none] (6) at (-0.5, 1.5) {};
		\node [style=none] (7) at (-0.5, 1.5) {};
		\node [style=none] (8) at (-0.5, 1.5) {};
		\node [style=none] (9) at (-2, 1.5) {};
		\node [style=none] (10) at (-2, 1.5) {};
		\node [style=none] (11) at (-2, 1.5) {};
		\node [style=none] (12) at (-1.25, -0.5) {};
		\node [style=none] (13) at (-1.25, 0.25) {};
		\node [style=none] (14) at (-1.25, 0.25) {};
		\node [style=white dot] (15) at (-1.25, 0.25) {};
		\node [style=none] (16) at (-1.25, 0.25) {};
		\node [style=none] (17) at (1.25, 0.25) {};
		\node [style=none] (18) at (2, 1.5) {};
		\node [style=none] (19) at (1.25, -0.5) {};
		\node [style=none] (20) at (0.5, 1.5) {};
		\node [style=none] (21) at (2, 1.5) {};
		\node [style=none] (22) at (0.5, 1.5) {};
		\node [style=none] (23) at (2, 1.5) {};
		\node [style=none] (24) at (1.25, 0.25) {};
		\node [style=none] (25) at (1.25, 0.25) {};
		\node [style=none] (26) at (0.5, 1.5) {};
		\node [style=none] (27) at (2, 1.5) {};
		\node [style=none] (28) at (2, 1.5) {};
		\node [style=none] (29) at (0.5, 1.5) {};
		\node [style=white dot] (30) at (1.25, 0.25) {};
		\node [style=none] (31) at (0.5, 1.5) {};
		\node [style=none] (32) at (2, 1.5) {};
		\node [style=kpoint] (33) at (1.25, -0.75) {\footnotesize$verb_r$};
	\end{pgfonlayer}
	\begin{pgfonlayer}{edgelayer}
		\draw [style=swap] (12.center) to (14.center);
		\draw [style=swap, in=165, out=-90, looseness=1.25] (11.center) to (13.center);
		\draw [style=swap, in=21, out=-90, looseness=1.25] (8.center) to (13.center);
		\draw [style=swap] (19.center) to (17.center);
		\draw [style=swap, in=165, out=-90, looseness=1.25] (22.center) to (25.center);
		\draw [style=swap, in=21, out=-90, looseness=1.25] (21.center) to (25.center);
	\end{pgfonlayer}
\end{tikzpicture}\end{gathered}
 \]
 
Secondly, a verb can be built by copying just one dimension, giving a sentence space $N$. Again, the verb tensor is approximately separable, giving the following diagrams:

\[
\begin{gathered}\begin{tikzpicture}
	\begin{pgfonlayer}{nodelayer}
		\node [style=none] (0) at (-2, 1.5) {};
		\node [style=none] (1) at (-0.5, 1.5) {};
		\node [style=none] (2) at (-2, 1.5) {};
		\node [style=none] (3) at (-0.5, 1.5) {};
		\node [style=none] (4) at (-0.5, 1.5) {};
		\node [style=none] (5) at (-0.5, 1.5) {};
		\node [style=none] (6) at (-0.5, 1.5) {};
		\node [style=none] (7) at (-0.5, 1.5) {};
		\node [style=none] (8) at (-2, 1.5) {};
		\node [style=none] (9) at (-2, 1.5) {};
		\node [style=none] (10) at (-2, 1.5) {};
		\node [style=none] (11) at (-1.25, -0.5) {};
		\node [style=none] (12) at (-1.25, 0.25) {};
		\node [style=none] (13) at (-1.25, 0.25) {};
		\node [style=white dot] (14) at (-1.25, 0.25) {};
		\node [style=none] (15) at (-1.25, 0.25) {};
		\node [style=none] (16) at (1.25, 1.5) {};
		\node [style=none] (17) at (1.25, -0.5) {};
		\node [style=none] (18) at (1.25, 1.5) {};
		\node [style=none] (19) at (1.25, 1.5) {};
		\node [style=none] (20) at (1.25, 1.5) {};
		\node [style=none] (21) at (1.25, 1.5) {};
		\node [style=none] (22) at (1.25, 1.5) {};
		\node [style=kpoint, minimum width=1.3225 cm, inner sep=2 pt] (23) at (0, -0.75) {\footnotesize $verb$};
	\end{pgfonlayer}
	\begin{pgfonlayer}{edgelayer}
		\draw [style=swap] (11.center) to (13.center);
		\draw [style=swap, in=165, out=-90, looseness=1.25] (10.center) to (12.center);
		\draw [style=swap, in=21, out=-90, looseness=1.25] (7.center) to (12.center);
		\draw (16.center) to (17.center);
	\end{pgfonlayer}
\end{tikzpicture}\end{gathered} \quad
\approx \quad \begin{gathered}\begin{tikzpicture}
	\begin{pgfonlayer}{nodelayer}
		\node [style=none] (0) at (-2, 1.5) {};
		\node [style=none] (1) at (-0.5, 1.5) {};
		\node [style=none] (2) at (-2, 1.5) {};
		\node [style=none] (3) at (-0.5, 1.5) {};
		\node [style=none] (4) at (-0.5, 1.5) {};
		\node [style=none] (5) at (-0.5, 1.5) {};
		\node [style=none] (6) at (-0.5, 1.5) {};
		\node [style=none] (7) at (-0.5, 1.5) {};
		\node [style=none] (8) at (-2, 1.5) {};
		\node [style=none] (9) at (-2, 1.5) {};
		\node [style=none] (10) at (-2, 1.5) {};
		\node [style=none] (11) at (-1.25, -0.5) {};
		\node [style=none] (12) at (-1.25, 0.25) {};
		\node [style=none] (13) at (-1.25, 0.25) {};
		\node [style=white dot] (14) at (-1.25, 0.25) {};
		\node [style=none] (15) at (-1.25, 0.25) {};
		\node [style=none] (16) at (1.25, 1.5) {};
		\node [style=none] (17) at (1.25, -0.5) {};
		\node [style=none] (18) at (1.25, 1.5) {};
		\node [style=none] (19) at (1.25, 1.5) {};
		\node [style=none] (20) at (1.25, 1.5) {};
		\node [style=none] (21) at (1.25, 1.5) {};
		\node [style=none] (22) at (1.25, 1.5) {};
		\node [style=kpoint] (23) at (-1.25, -0.75) {\footnotesize $verb_l$};
		\node [style=kpoint] (24) at (1.25, -0.75) {\footnotesize $verb_r$};
	\end{pgfonlayer}
	\begin{pgfonlayer}{edgelayer}
		\draw [style=swap] (11.center) to (13.center);
		\draw [style=swap, in=165, out=-90, looseness=1.25] (10.center) to (12.center);
		\draw [style=swap, in=21, out=-90, looseness=1.25] (7.center) to (12.center);
		\draw (16.center) to (17.center);
	\end{pgfonlayer}
\end{tikzpicture}\end{gathered}, \qquad
\begin{gathered}\begin{tikzpicture}
	\begin{pgfonlayer}{nodelayer}
		\node [style=none] (0) at (0.5, 1.5) {};
		\node [style=none] (1) at (2, 1.5) {};
		\node [style=none] (2) at (0.5, 1.5) {};
		\node [style=none] (3) at (2, 1.5) {};
		\node [style=none] (4) at (2, 1.5) {};
		\node [style=none] (5) at (2, 1.5) {};
		\node [style=none] (6) at (2, 1.5) {};
		\node [style=none] (7) at (2, 1.5) {};
		\node [style=none] (8) at (0.5, 1.5) {};
		\node [style=none] (9) at (0.5, 1.5) {};
		\node [style=none] (10) at (0.5, 1.5) {};
		\node [style=none] (11) at (1.25, -0.5) {};
		\node [style=none] (12) at (1.25, 0.25) {};
		\node [style=none] (13) at (1.25, 0.25) {};
		\node [style=white dot] (14) at (1.25, 0.25) {};
		\node [style=none] (15) at (1.25, 0.25) {};
		\node [style=none] (16) at (-1.25, 1.5) {};
		\node [style=none] (17) at (-1.25, -0.5) {};
		\node [style=none] (18) at (-1.25, 1.5) {};
		\node [style=none] (19) at (-1.25, 1.5) {};
		\node [style=none] (20) at (-1.25, 1.5) {};
		\node [style=none] (21) at (-1.25, 1.5) {};
		\node [style=none] (22) at (-1.25, 1.5) {};
		\node [style=kpoint, minimum width=1.3225 cm, inner sep=2 pt] (23) at (0, -0.75) {\footnotesize $verb$};
	\end{pgfonlayer}
	\begin{pgfonlayer}{edgelayer}
		\draw [style=swap] (11.center) to (13.center);
		\draw [style=swap, in=165, out=-90, looseness=1.25] (10.center) to (12.center);
		\draw [style=swap, in=21, out=-90, looseness=1.25] (7.center) to (12.center);
		\draw (16.center) to (17.center);
	\end{pgfonlayer}
\end{tikzpicture}\end{gathered} \quad
\approx \quad \begin{gathered}\begin{tikzpicture}
	\begin{pgfonlayer}{nodelayer}
		\node [style=none] (0) at (0.5, 1.5) {};
		\node [style=none] (1) at (2, 1.5) {};
		\node [style=none] (2) at (0.5, 1.5) {};
		\node [style=none] (3) at (2, 1.5) {};
		\node [style=none] (4) at (2, 1.5) {};
		\node [style=none] (5) at (2, 1.5) {};
		\node [style=none] (6) at (2, 1.5) {};
		\node [style=none] (7) at (2, 1.5) {};
		\node [style=none] (8) at (0.5, 1.5) {};
		\node [style=none] (9) at (0.5, 1.5) {};
		\node [style=none] (10) at (0.5, 1.5) {};
		\node [style=none] (11) at (1.25, -0.5) {};
		\node [style=none] (12) at (1.25, 0.25) {};
		\node [style=none] (13) at (1.25, 0.25) {};
		\node [style=white dot] (14) at (1.25, 0.25) {};
		\node [style=none] (15) at (1.25, 0.25) {};
		\node [style=none] (16) at (-1.25, 1.5) {};
		\node [style=none] (17) at (-1.25, -0.5) {};
		\node [style=none] (18) at (-1.25, 1.5) {};
		\node [style=none] (19) at (-1.25, 1.5) {};
		\node [style=none] (20) at (-1.25, 1.5) {};
		\node [style=none] (21) at (-1.25, 1.5) {};
		\node [style=none] (22) at (-1.25, 1.5) {};
		\node [style=kpoint] (23) at (-1.25, -0.75) {\footnotesize $verb_l$};
		\node [style=kpoint] (24) at (1.25, -0.75) {\footnotesize $verb_r$};
	\end{pgfonlayer}
	\begin{pgfonlayer}{edgelayer}
		\draw [style=swap] (11.center) to (13.center);
		\draw [style=swap, in=165, out=-90, looseness=1.25] (10.center) to (12.center);
		\draw [style=swap, in=21, out=-90, looseness=1.25] (7.center) to (12.center);
		\draw (16.center) to (17.center);
	\end{pgfonlayer}
\end{tikzpicture}\end{gathered}
 \]
 
In these last two cases, when combined with nouns, the sentence reduces to an intersective interaction between either the subject and the verb, or the object and the verb. In the object case:
 \[
\begin{gathered}\begin{tikzpicture}
	\begin{pgfonlayer}{nodelayer}
		\node [style=none] (0) at (2, 0.75) {};
		\node [style=none] (1) at (2, 0.75) {};
		\node [style=none] (2) at (2, 0.75) {};
		\node [style=none] (3) at (2, 0.75) {};
		\node [style=none] (4) at (2, 0.75) {};
		\node [style=none] (5) at (0, 2.5) {};
		\node [style=none] (6) at (1, -1) {};
		\node [style=none] (7) at (1, 0) {};
		\node [style=none] (8) at (1, 0) {};
		\node [style=white dot] (9) at (1, 0) {};
		\node [style=none] (10) at (-2, 0.75) {};
		\node [style=none] (11) at (-2, -1) {};
		\node [style=kpoint] (12) at (-2, -1.25) {\footnotesize $hates_l$};
		\node [style=kpoint] (13) at (1, -1.25) {\footnotesize $hates_r$};
		\node [style=kpoint] (14) at (-4.5, -1.25) {\footnotesize $A$};
		\node [style=none] (15) at (-4.5, -1) {};
		\node [style=none] (16) at (-4.5, 0.75) {};
		\node [style=kpoint] (17) at (3.75, -1.25) {\footnotesize $B$};
		\node [style=none] (18) at (3.75, 0.75) {};
		\node [style=none] (19) at (3.75, 0.75) {};
		\node [style=none] (20) at (3.75, -1) {};
		\node [style=none] (21) at (3.75, 0.75) {};
		\node [style=none] (22) at (3.75, 0.75) {};
		\node [style=none] (23) at (3.75, 0.75) {};
		\node [style=none] (24) at (3.75, 0.75) {};
	\end{pgfonlayer}
	\begin{pgfonlayer}{edgelayer}
		\draw [style=swap] (6.center) to (8.center);
		\draw [style=swap, in=180, out=-90, looseness=1.00] (5.center) to (7.center);
		\draw [style=swap, in=0, out=-90, looseness=1.25] (4.center) to (7.center);
		\draw (10.center) to (11.center);
		\draw (16.center) to (15.center);
		\draw (24.center) to (20.center);
		\draw [bend left=90, looseness=1.25] (16.center) to (10.center);
		\draw [bend left=90, looseness=1.75] (4.center) to (18.center);
	\end{pgfonlayer}
\end{tikzpicture}\end{gathered} \quad
\approx \quad \begin{gathered}\begin{tikzpicture}
	\begin{pgfonlayer}{nodelayer}
		\node [style=none] (0) at (1, 2.5) {};
		\node [style=none] (1) at (-0.5, -1) {};
		\node [style=none] (2) at (1, 0) {};
		\node [style=white dot] (3) at (1, 0) {};
		\node [style=kpoint] (4) at (-0.5, -1.25) {\footnotesize $hates_r$};
		\node [style=kpoint] (5) at (2.5, -1.25) {\footnotesize $B$};
		\node [style=none] (6) at (2.5, -1) {};
		\node [style=none] (7) at (1, 0) {};
	\end{pgfonlayer}
	\begin{pgfonlayer}{edgelayer}
		\draw [style=swap, in=180, out=90, looseness=1.00] (1.center) to (2.center);
		\draw [in=90, out=0, looseness=1.00] (7.center) to (6.center);
		\draw (0.center) to (2.center);
	\end{pgfonlayer}
\end{tikzpicture}\end{gathered}
 \]
Therefore, \cite{KartsaklisSadrzadeh2014} give a number of proposals for combining these two parts: namely taking the tensor product of the two parts, taking the Frobenius product of the two parts, or adding them. The best performance is obtained by the additive model. In what follows, we introduce the concept of `Cartesian' verbs, which are separable in the same sense as above.
 
\section{Cartesian verbs}
Just as  intersective adjectives have a reduced  meaning space (i.e.~from two wires to one wire), we will now characterise a kind of transitive verbs where something similar happens.  Even though this is an idealisation of general transitive verbs, it may still provide a useful practical  approximation of most verbs.  Moreover, the notion  encompasses most prepositions, which can be thought of as transitive verbs simply by adjoining ``being'' in front of them e.g.~``being on'', ``being next to'', ``being after'' etc.

Consider the sentence: ``flowerpot (is on) table''.  Probably when you just read this sentence you, just like us,  imagined some flowerpot on some table: 
\[
\includegraphics[height=3cm]{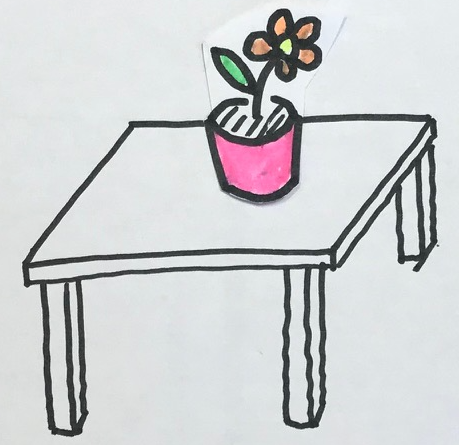}
\]
There is no reason not to take that image of a flowerpot on a table to be the representation of the meaning of the sentence ``flowerpot (is on) table''.  
 
Now, how is this image formed?  It consists of a picture of a flowerpot, one of a table, and the additional piece of data that the flowerpot is `on' the table: 
\[
\raisebox{-0.6cm}{\includegraphics[height=1.5cm]{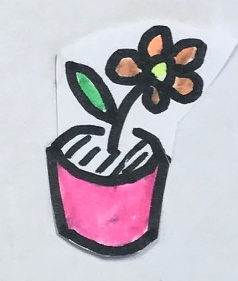}} 
\ \ \mbox{ on } \ \ 
\raisebox{-0.9cm}{\includegraphics[height=2.2cm]{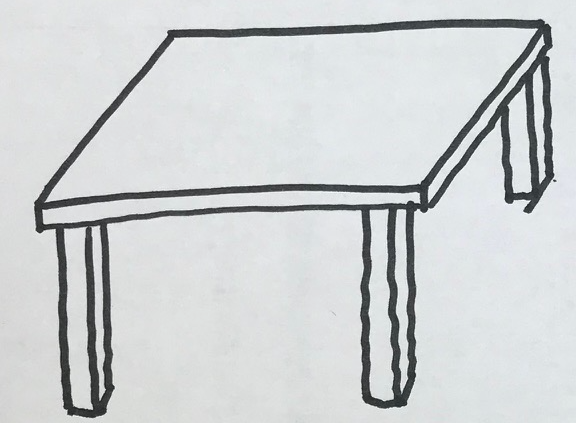}}
\ \  = \ \ 
\raisebox{-1.4cm}{\includegraphics[height=3cm]{figures/flowerpotontable}}
\]
There are two very important points to make here: 
\begin{enumerate}
\item \underline{Subject and object are unaltered} when interacting with the verb  i.e.~the images of the flowerpot and of the table stay exactly the same before and after we have put the flowerpot on the table.
\item \underline{A new feature (NF) arises} by bringing object and subject together, and which is specified by the verb i.e.~once we have two things in physical space, the table and the flowerpot, a new feature arises that made no sense before, namely, their relative location.
\end{enumerate}
This indicates the following internal wiring of ``being on'':
\ctikzfig{new1}
This structure indeed leaves object and subject unaltered when bringing them together in the final image, but also adjoins the new feature of relative position.  The data resulting from a sentence is then indeed the two images together with the specification of their relative position: 
\[
\begin{tikzpicture}
	\begin{pgfonlayer}{nodelayer}
		\node [style=none] (0) at (-3, 0.75) {};
		\node [style=none] (1) at (3, 0.75) {};
		\node [style=none] (2) at (1, 0.75) {};
		\node [style=kpoint] (3) at (0, -0.75) {\footnotesize``on"};
		\node [style=none] (4) at (-1, 0.75) {};
		\node [style=none] (5) at (0, -0.5) {};
		\node [style=none] (6) at (1, 0.75) {};
		\node [style=none] (7) at (-3, 0.75) {};
		\node [style=none] (8) at (-1, 0.75) {};
		\node [style=none] (9) at (-1, 0.75) {};
		\node [style=none] (10) at (0, 0.75) {};
		\node [style=none] (11) at (1, 0.75) {};
		\node [style=none] (12) at (3, 0.75) {};
		\node [style=none] (13) at (3, 0.75) {};
		\node [style=none] (14) at (1, 0.75) {};
		\node [style=none] (15) at (1, 0.75) {};
		\node [style=none] (16) at (1, 0.75) {};
		\node [style=none] (17) at (-1, 0.75) {};
		\node [style=none] (18) at (-1, 0.75) {};
		\node [style=none] (19) at (-1, 0.75) {};
		\node [style=none] (20) at (-3, 0.75) {};
		\node [style=none] (21) at (-3, 0.75) {};
		\node [style=none] (22) at (-3, 0.75) {};
		\node [style=kpoint] (23) at (-5, -0.75) {\footnotesize Sub};
		\node [style=kpoint] (24) at (5, -0.75) {\footnotesize Obj};
		\node [style=none] (25) at (5, 0.75) {};
		\node [style=none] (26) at (-5, 0.75) {};
		\node [style=none] (27) at (-1, 1.5) {};
		\node [style=none] (28) at (1, 1.5) {};
		\node [style=none] (29) at (1, 1.5) {};
		\node [style=none] (30) at (1, 1.5) {};
		\node [style=none] (31) at (-1, 1.5) {};
		\node [style=none] (32) at (0, 1.5) {};
		\node [style=none] (33) at (-1, 1.5) {};
		\node [style=none] (34) at (-1, 1.5) {};
		\node [style=none] (35) at (-1, 1.5) {};
		\node [style=none] (36) at (1, 1.5) {};
		\node [style=none] (37) at (-1, 1.5) {};
		\node [style=none] (38) at (1, 1.5) {};
		\node [style=none] (39) at (1, 1.5) {};
		\node [style=none] (40) at (3.5, -1.75) {};
		\node [style=none] (41) at (-3.5, 0.75) {};
		\node [style=none] (42) at (-3.5, -1.75) {};
		\node [style=none] (43) at (3.5, 0.75) {};
	\end{pgfonlayer}
	\begin{pgfonlayer}{edgelayer}
		\draw [style=swap] (5.center) to (10.center);
		\draw [style=swap, bend right=90, looseness=1.75] (7.center) to (8.center);
		\draw [style=swap, bend right=90, looseness=1.75] (6.center) to (12.center);
		\draw [style=swap, bend left=90, looseness=1.50] (26.center) to (22.center);
		\draw [style=swap, bend left=90, looseness=1.50] (13.center) to (25.center);
		\draw [style=swap] (25.center) to (24);
		\draw [style=swap] (26.center) to (23);
		\draw [style=swap] (32.center) to (10.center);
		\draw [style=swap] (28.center) to (16.center);
		\draw [style=swap] (27.center) to (19.center);
		\draw [style=dashed] (41.center) to (42.center);
		\draw [style=dashed] (42.center) to (40.center);
		\draw [style=dashed] (43.center) to (40.center);
		\draw [style=dashed] (41.center) to (43.center);
	\end{pgfonlayer}
\end{tikzpicture}\ \  = \ \ \begin{tikzpicture}
	\begin{pgfonlayer}{nodelayer}
		\node [style=none] (0) at (2.25, 0.75) {};
		\node [style=kpoint] (1) at (0, -0.5) {\footnotesize``on"};
		\node [style=none] (2) at (-2.25, 0.75) {};
		\node [style=none] (3) at (0, -0.25) {};
		\node [style=none] (4) at (2.25, 0.75) {};
		\node [style=none] (5) at (-2.25, 0.75) {};
		\node [style=none] (6) at (-2.25, 0.75) {};
		\node [style=none] (7) at (0, 0.75) {};
		\node [style=none] (8) at (2.25, 0.75) {};
		\node [style=none] (9) at (2.25, 0.75) {};
		\node [style=none] (10) at (2.25, 0.75) {};
		\node [style=none] (11) at (2.25, 0.75) {};
		\node [style=none] (12) at (-2.25, 0.75) {};
		\node [style=none] (13) at (-2.25, 0.75) {};
		\node [style=none] (14) at (-2.25, 0.75) {};
		\node [style=kpoint] (15) at (-2.25, -0.5) {\footnotesize Sub};
		\node [style=kpoint] (16) at (2.25, -0.5) {\footnotesize Obj};
		\node [style=none] (17) at (2.25, 0.75) {};
		\node [style=none] (18) at (-2.25, 0.75) {};
	\end{pgfonlayer}
	\begin{pgfonlayer}{edgelayer}
		\draw [style=swap] (3.center) to (7.center);
		\draw [style=swap] (17.center) to (16);
		\draw [style=swap] (18.center) to (15);
	\end{pgfonlayer}
\end{tikzpicture}
\]
We see that we now have three disconnected triangles, the object, the subject, and one representing the verb.  We refer to such a situation where the whole is described by its parts as \em Cartesian\em.  (this indeed corresponds to the description of the states in a Cartesian category) However, do note that the verb itself is clearly does not separate into triangles of the ${}^{-1}n$, the $s$ and the $n^{-1}$ type (witnessed by the cups connecting these), since this would mean that the meaning of the sentence does not depend on its object and its subject.
 
\begin{defn}
\label{def:Cartesian}
A transitive verb is called \em Cartesian \em if it has the following internal wiring: 
\ctikzfig{new4}
\end{defn}

The subject  and object not being altered by the verb is a strong constraint, since in most cases the verb would provide additional specification about the subject and the object.  The tamest case of this is that the action of the verb on subject  and on the object takes the form of an intersective adjective, possibly modified in an `active' and `passive' sense respectively.  For example, Bob waters flowers modifies Bob in that we now know that he has a device on him to water plants, and it modifies the plant in that we know that its soil will be  wet.  Both of these are clearly intersective modifications. 

\begin{defn}
\label{def:semicart}
A transitive verb is called \em semi-Cartesian \em if it has the following internal wiring: 
\ctikzfig{new5}
where  P${}_s$ and P${}_o$ represent the adjective effect of the verb on subject and object respectively.
\end{defn}

An example of such a  semi-Cartesian verb is ``paint red", which adjoins a brush to the subject who is painting and colours the object being painted: 
\ctikzfig{new7} 
For a general semi-Cartesian transitive verb sentence the meaning is now as follows:   
\ctikzfig{new6}

The action of a semi-Cartesian transitive verb sentence can be seen as `semantically type checking' its subject and object. The intersective action of the subject and object parts of the verb mean that if the subject or object is of an inappropriate type, the sentence has no meaning.

\begin{example}[Semantic type-checking]
Consider the categorical compositional model applied to the semantic category $\ConvexRel$, introduced in \cite{ConcSpacI}. $\ConvexRel$ is a category used to describe conceptual spaces \cite{gardenfors}. We specify a meaning space $N$ as $N_{\lang{colour}} \otimes N_{\lang{taste}} \otimes N_{\lang{texture}}$,
and we describe some nouns pictorially as:

\begin{align*}
\lang{banana} \ \ &=\ \  \begin{gathered} \includegraphics[width=0.15\textwidth]{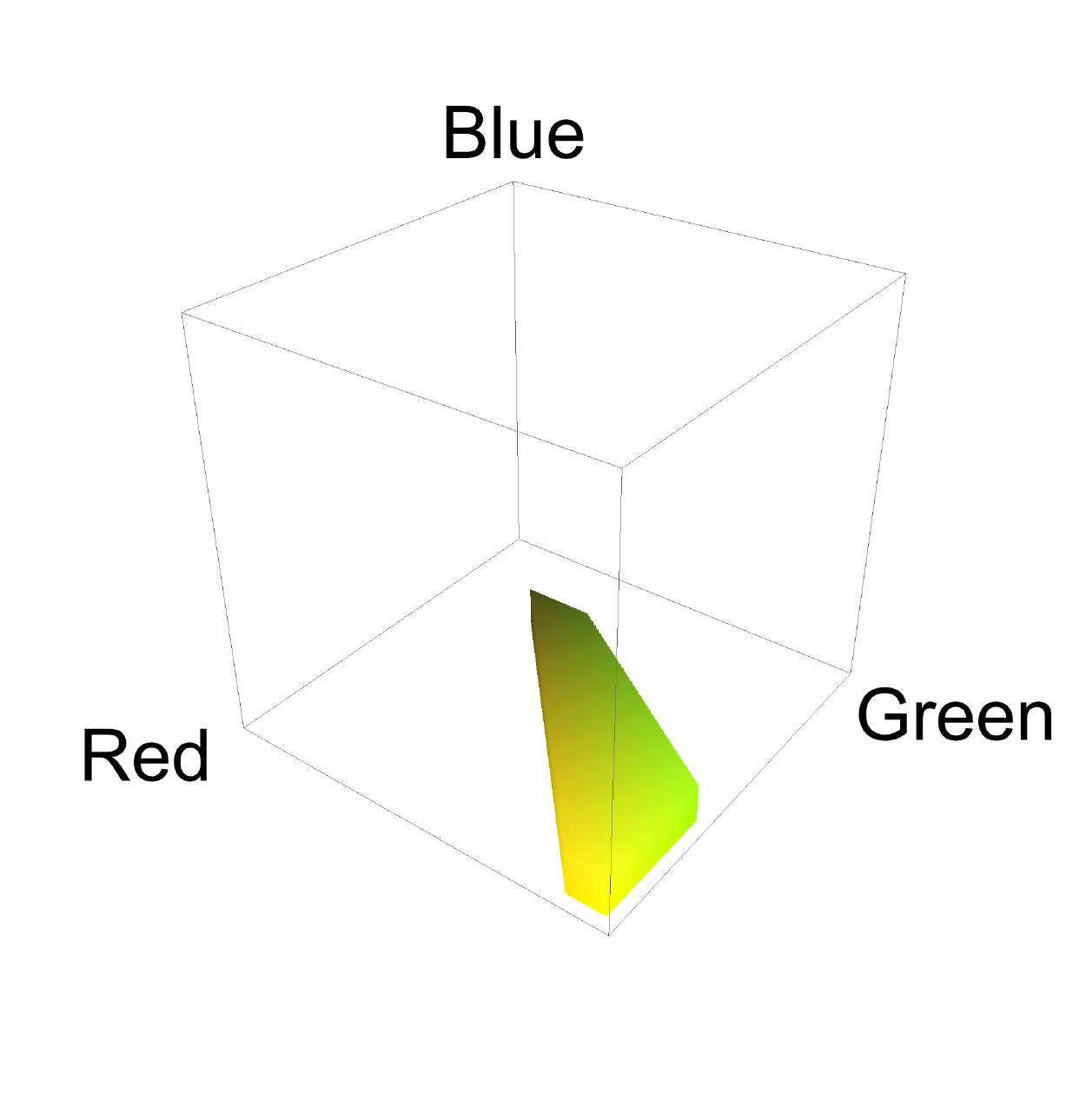}\end{gathered} \ \ \otimes\ \  \begin{gathered} \includegraphics[width=0.15\textwidth]{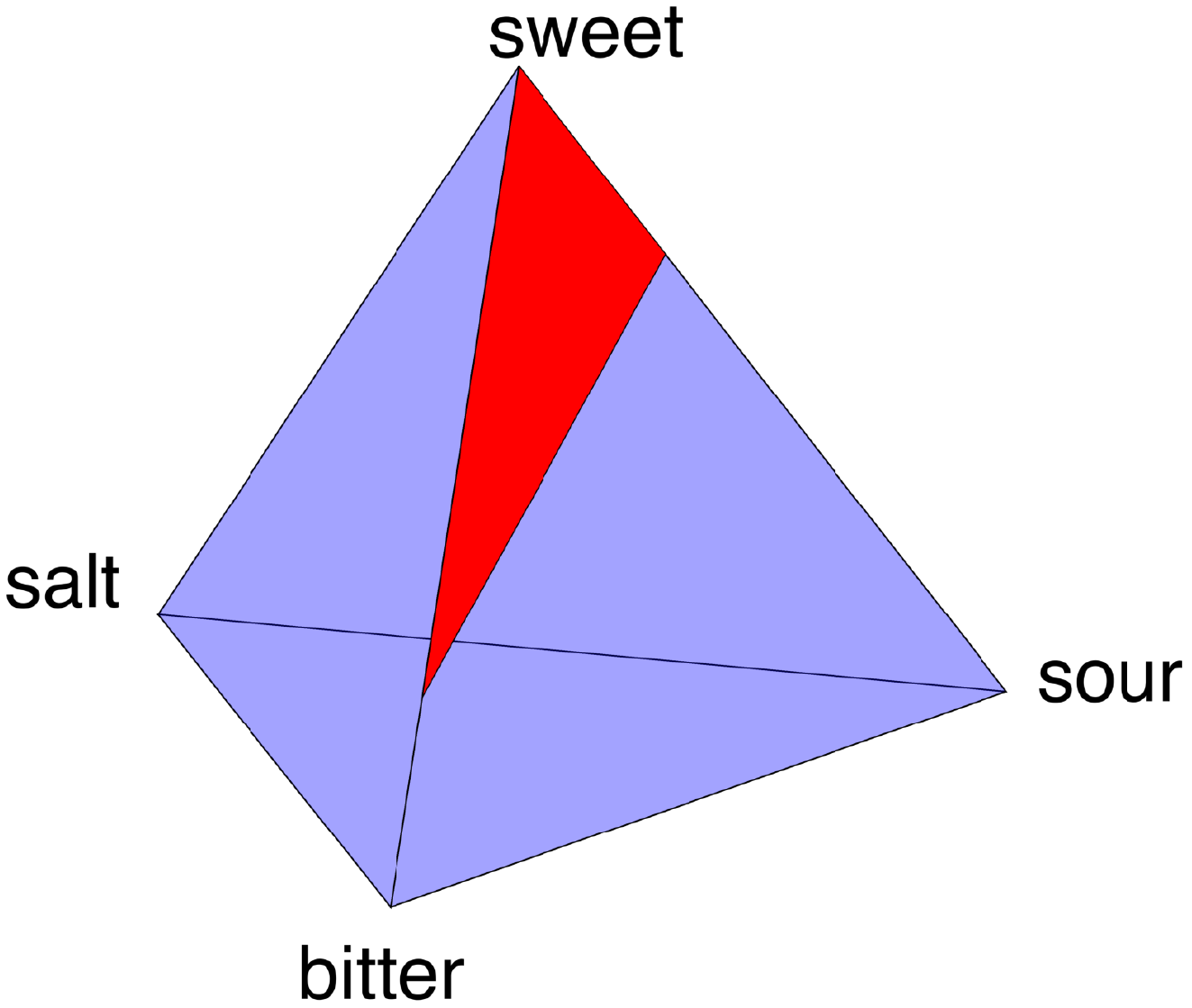}\end{gathered}\ \ \otimes\ \ \begin{gathered} \includegraphics[width=0.2\textwidth]{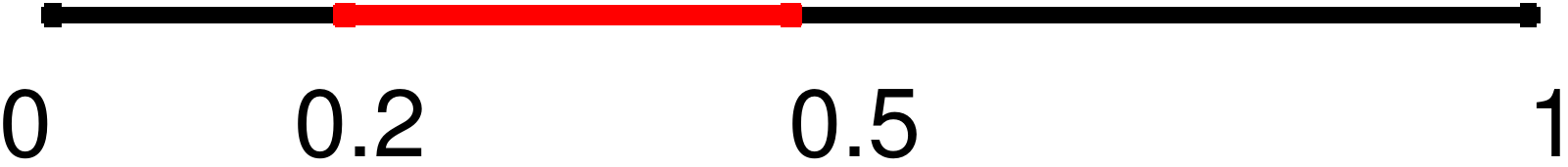}\end{gathered}\\
\lang{apple} \ \ &=\ \  \begin{gathered} \includegraphics[width=0.15\textwidth]{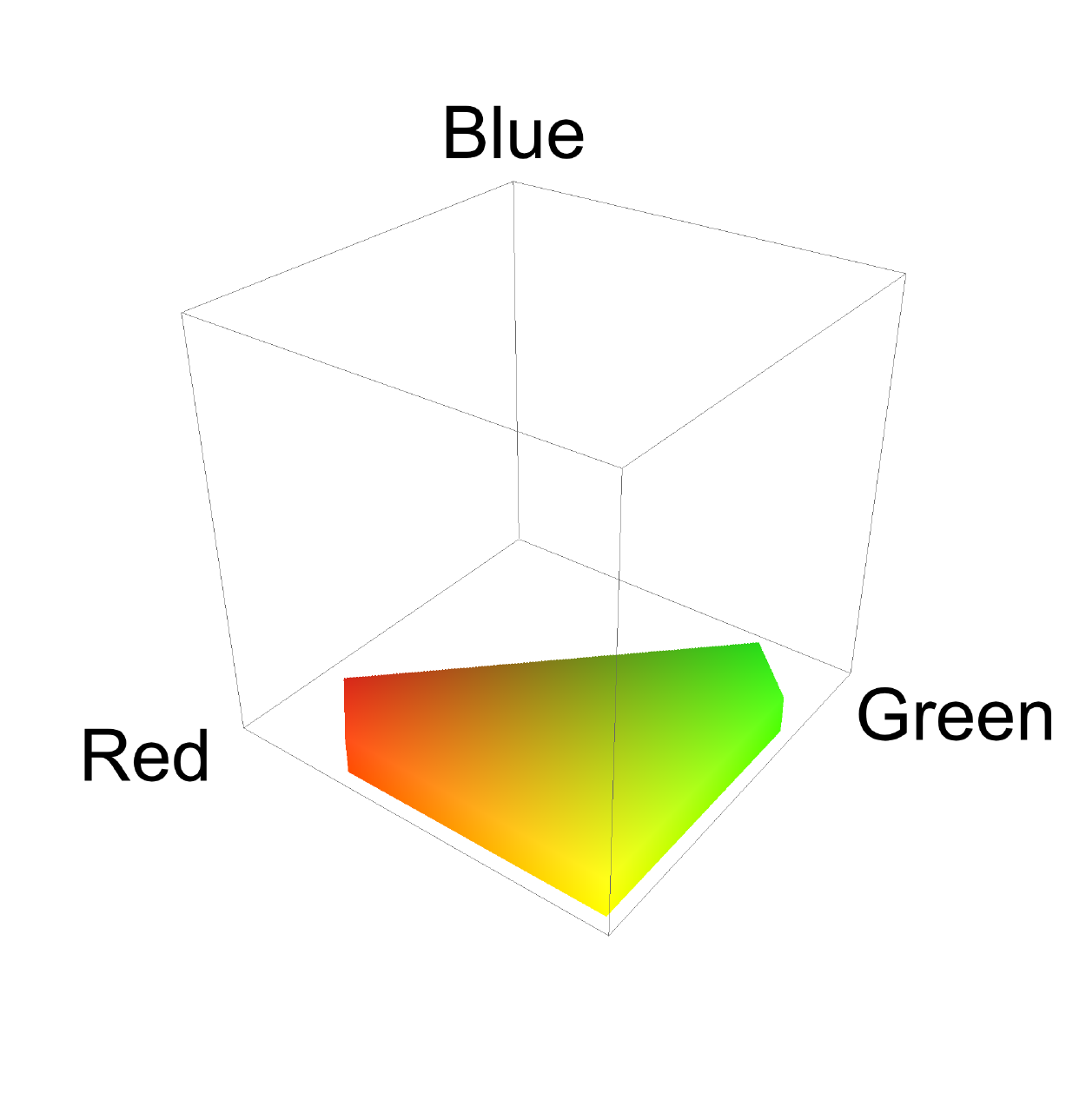}\end{gathered} \ \ \otimes\ \  \begin{gathered} \includegraphics[width=0.15\textwidth]{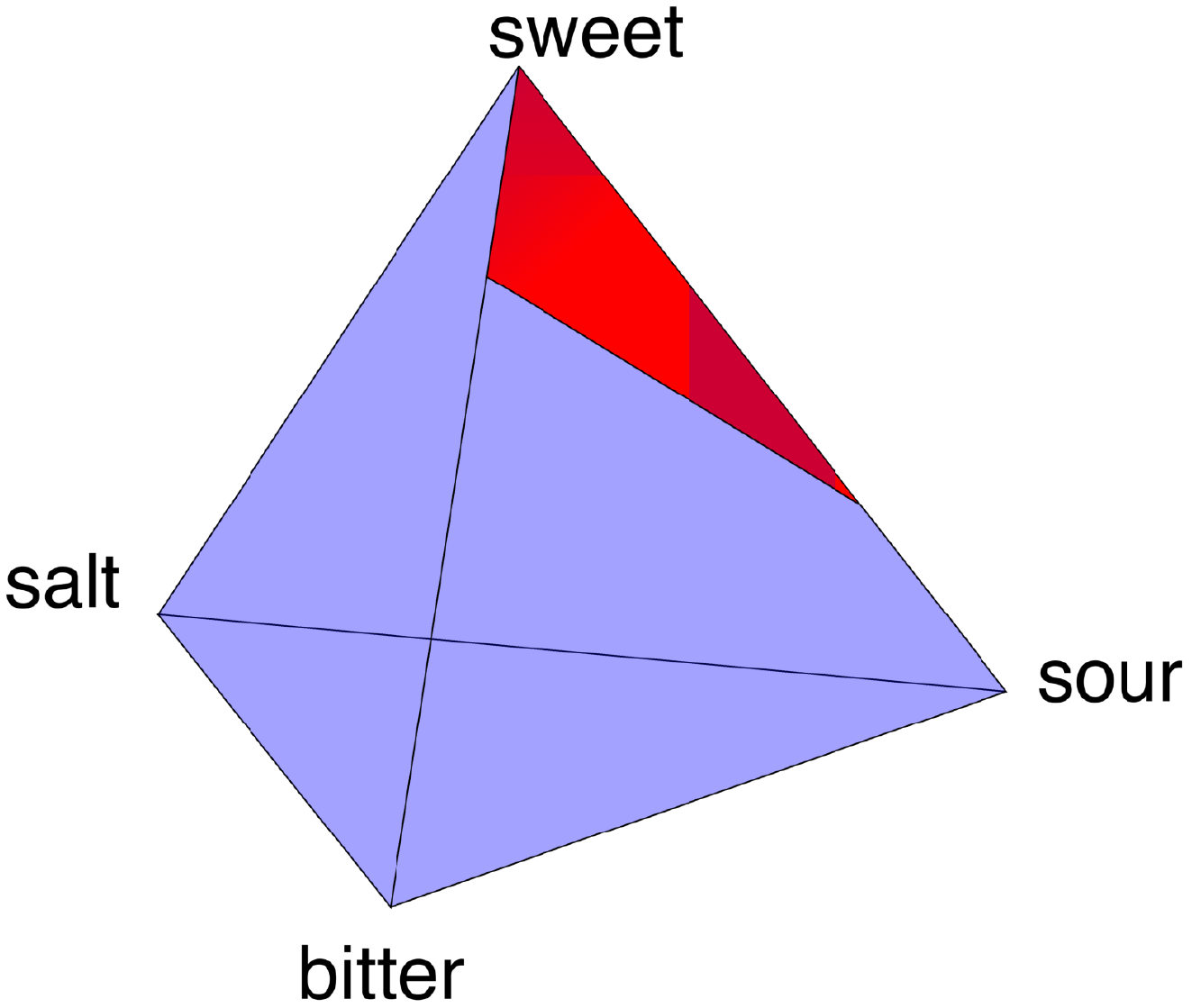}\end{gathered}\ \ \otimes\ \ \begin{gathered} \includegraphics[width=0.2\textwidth]{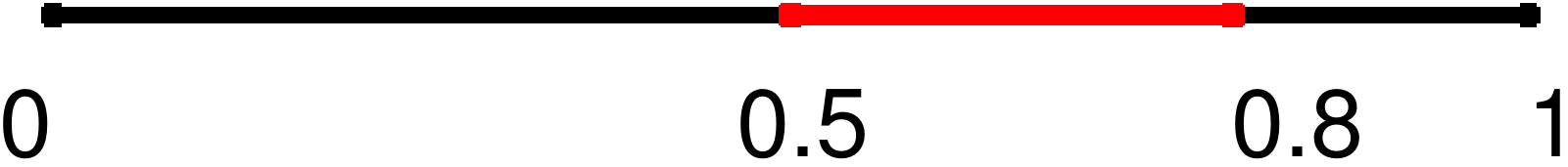}\end{gathered}\\
\lang{fire} \ \ &=\ \  \begin{gathered} \includegraphics[width=0.15\textwidth]{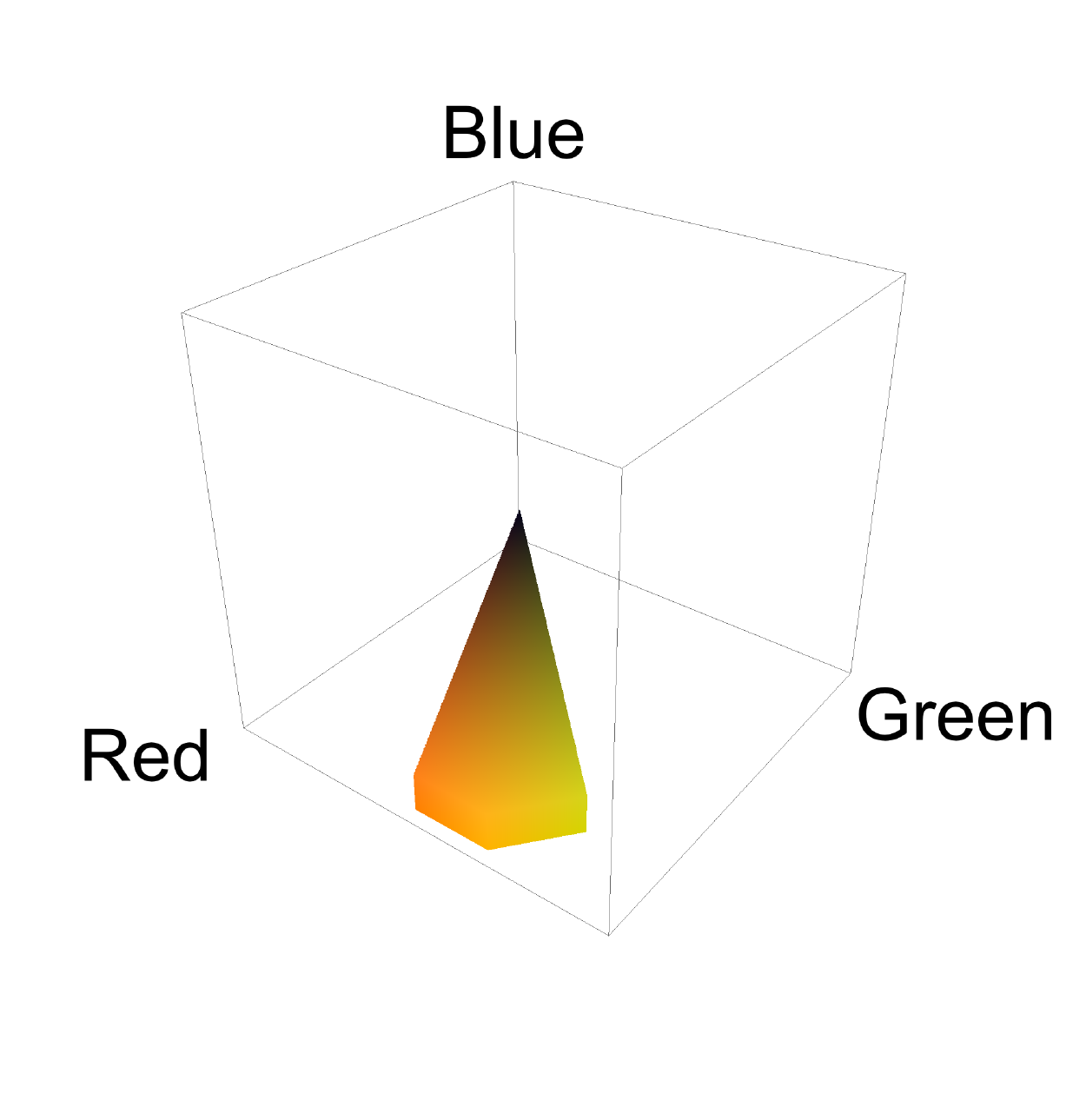}\end{gathered} \  \otimes\ \ \begin{gathered} \includegraphics[width=0.2\textwidth]{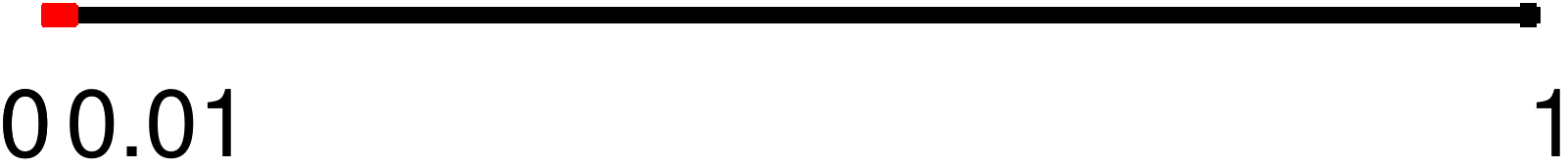}\end{gathered}\\
\end{align*}
Note that these can also all be specified equationally. 

Now, we can specify the verb $\lang{taste}$ as follows. Its subject can be any noun, but its object must  describe a flavour. So we specify it as being derived from one copy of the whole noun space, together with two copies of the taste space.
\[
\lang{taste} \ \  = \ \ 
\delta(N)  \ \ \otimes \ \  \begin{gathered} \includegraphics[width=0.15\textwidth]{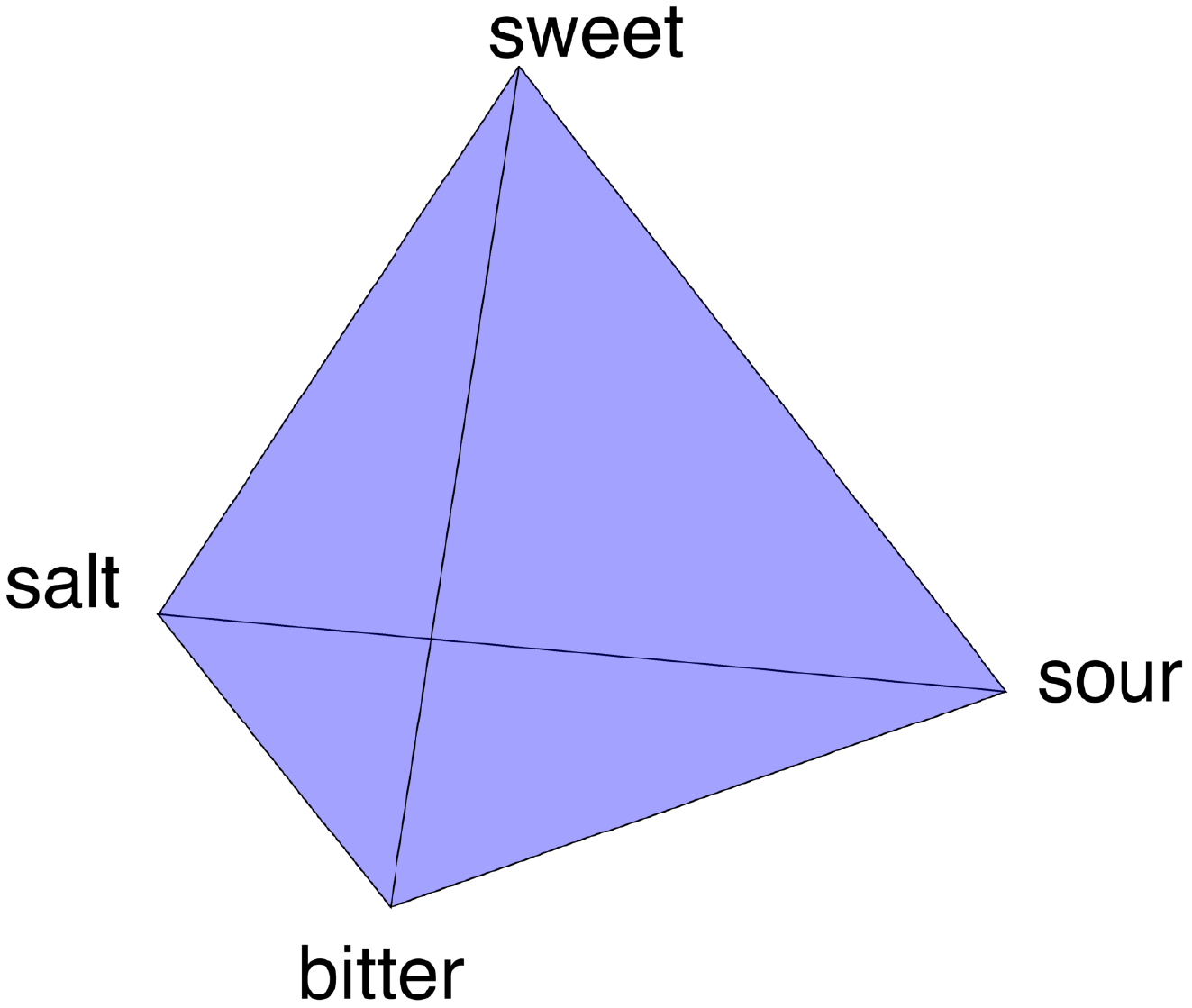}\end{gathered}  \ \ \otimes \ \  \delta\left(\begin{gathered} \includegraphics[width=0.15\textwidth]{pdfpics/taste.pdf}\end{gathered}\right)
\]
where $\delta$ is notation for the copying spider described in section \ref{sec:wirings}.
Then we have:
\begin{align*}
\lang{apple tastes sweet} &= 
\lang{apple} \otimes\ \ \begin{gathered} \includegraphics[width=0.15\textwidth]{pdfpics/taste.pdf}\end{gathered} \ \ \otimes\ \ \begin{gathered} \includegraphics[width=0.15\textwidth]{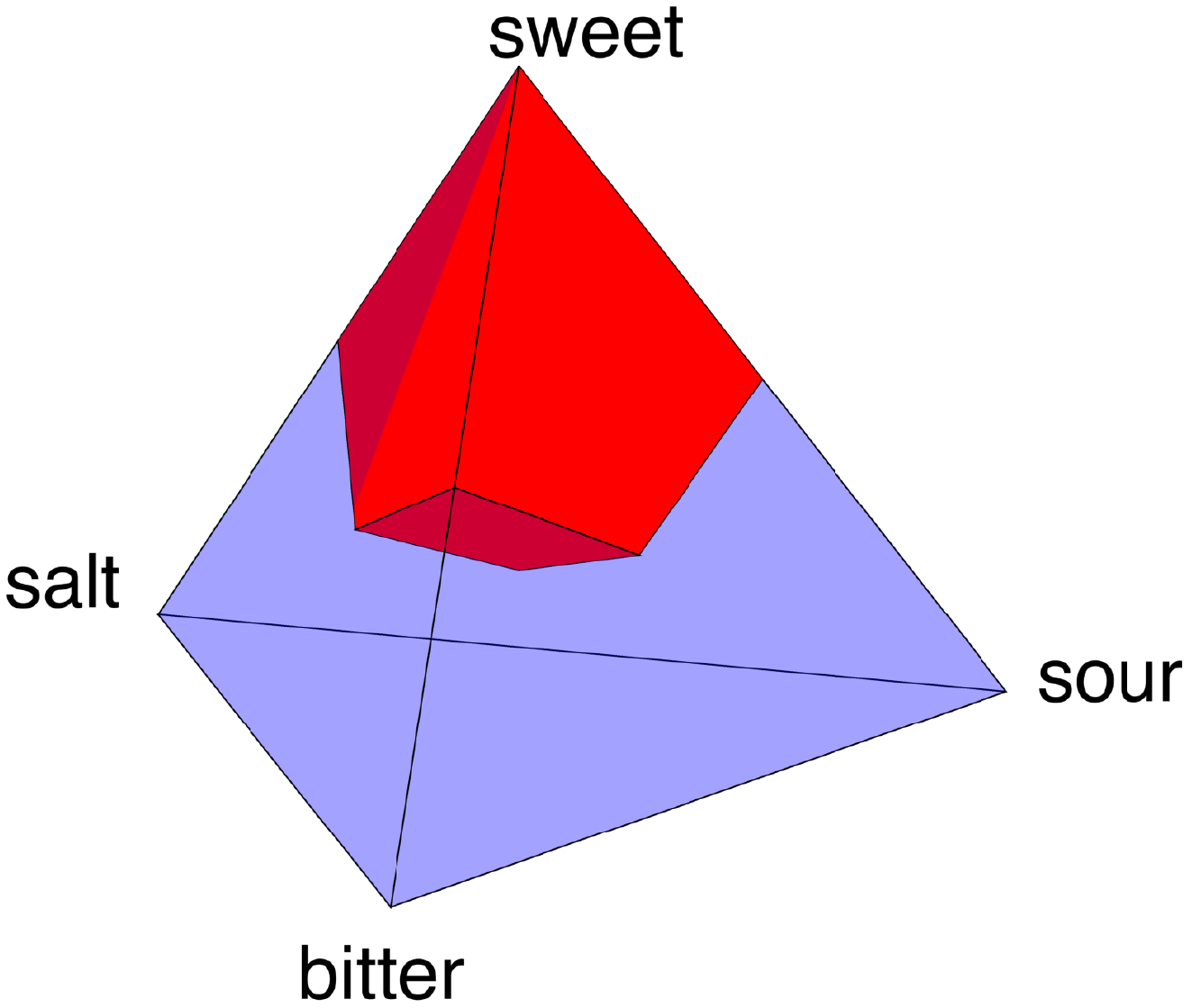}\end{gathered}
\end{align*}
and even:
\begin{align*}
\lang{apple tastes banana} &= 
\lang{apple} \otimes\ \ \begin{gathered} \includegraphics[width=0.15\textwidth]{pdfpics/taste.pdf}\end{gathered} \ \ \otimes\ \ \begin{gathered} \includegraphics[width=0.15\textwidth]{pdfpics/banana_taste_amended.pdf}\end{gathered}
\end{align*}
But the following sentence has no literal meaning:
\begin{align*}
\lang{apple tastes fire} &= 
\lang{apple} \otimes\ \ \begin{gathered} \includegraphics[width=0.15\textwidth]{pdfpics/taste.pdf}\end{gathered} \ \ \otimes\ \ \emptyset
=\emptyset
\end{align*}
\end{example}

In the above example, the new feature of the verb did not contribute much semantically, other than to emphasize that the verb was to do with taste. The following example uses a new feature that has more semantic import.

\begin{example}
We again work in $\ConvexRel$. We specify the meaning space to be
$N = N_{\lang{colour}} \otimes N_{\lang{taste}} \otimes N_{\lang{size}} \otimes N_{\lang{location}}$.
So suppose we have:
\begin{align*}
\lang{banana} \ \ &=\ \  \begin{gathered} \includegraphics[width=0.15\textwidth]{pdfpics/RGB_Banana.pdf}\end{gathered} \ \ \otimes\ \  \begin{gathered} \includegraphics[width=0.15\textwidth]{pdfpics/banana_taste_amended.pdf}\end{gathered}\ \ \otimes\ \ [15, 20] \ \ \otimes \mathbb{R}\\
\lang{apple} \ \ &=\ \  \begin{gathered} \includegraphics[width=0.15\textwidth]{pdfpics/RGB_Apple.pdf}\end{gathered} \ \ \otimes\ \  \begin{gathered} \includegraphics[width=0.15\textwidth]{pdfpics/apple_taste_amended.pdf}\end{gathered}\ \ \otimes\ \ [5, 10] \ \ \otimes \mathbb{R}\\
\end{align*}
and: 
\[
\lang{is next to} \ \  = \ \ 
\delta(N)  \ \ \otimes \ \  \begin{gathered} \includegraphics[width=0.2\textwidth]{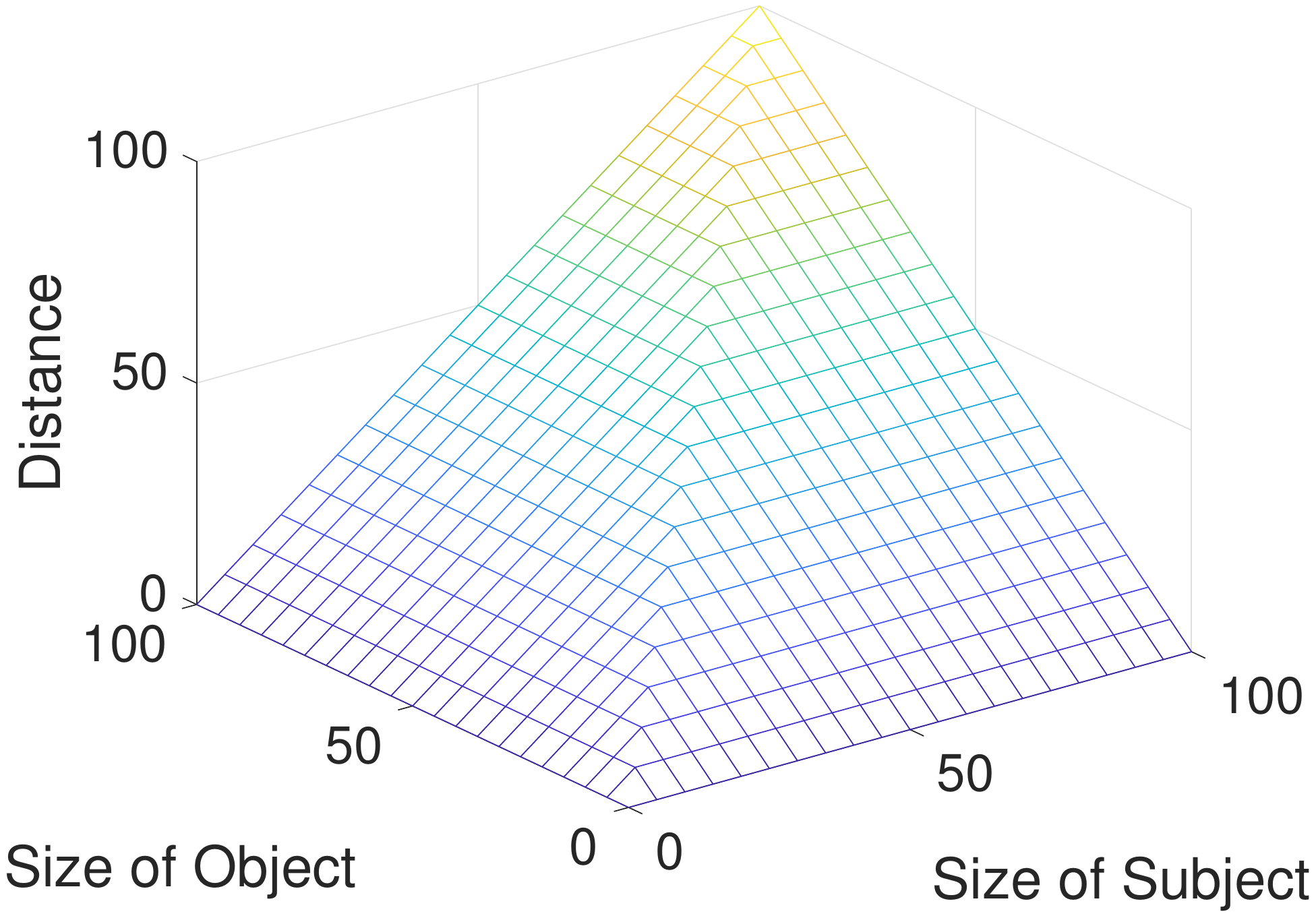}\end{gathered}  \ \ \otimes \ \  \delta(N)
\]
where the plot defines a relation between the sizes of the subject and the object of the verb and the distance between them, namely that the distance can be no greater than the size of the smaller of the pair. Then, the sentence $\lang{apple is next to banana}$ tells us that, if the sentence is true, the apple and the banana stand in that relationship to one another. It does not tell us what the locations of the apple and the banana are, and it does not tell us whether the sentence is true or not.
\end{example}


\section{Interaction with relative pronouns} 
Cartesian and semi-Cartesian verbs interact in an interesting ways with relative pronouns. In the analysis of~\cite{FrobMeanI, FrobMeanII}, the subject relative pronoun ``who'' has internal structure:
\ctikzfig{new8}
The structure of semi-Cartesian verbs suggests what the internal wiring of the spider that kills the sentence type should be.
For semi-Cartesian verbs, the type $s$ consists of three wires, of which the outer ones are both $n$ and the inner-wire which represents the new feature introduced by the verb is left unspecified.  We will see that assigning this central wire to have composite structure ${}^{-1}nn^{-1}$ allows us to to realize the intended behaviour of a relative pronoun,  with the ``sentence killing'' component chosen to be:
\ctikzfig{new9}
Note here that the particular connectedness indicates here that while the verb type consists of three separate wires, the middle one indicates how the two other ones, representing subject and object respectively, are in fact related. Plugging everything together in a sentence with a relative pronoun we obtain:
\ctikzfig{new10}
Notice how the double wire encoding the verb action allows information to flow as one might expect, by interacting appropriately with the ``sentence killing'' component.
Rearranging, this becomes:
\ctikzfig{new11}
where we used the key property of multi-wires, namely that they can fuse together, or de-fuse.
Taking the subject to be ``she", the verb to be ``paint red" and the object to be ``table", this indeed provides us with a conjunction of, a female person with a paintbrush with is with a table turning red, so the meaning of the noun-phrase is whoever that person is.

The observations in this section point to internal structure on our choice of sentence space. A detailed study of this phenomenon is left to later work.

\section{Empirical testing}
We undertake some simple empirical testing of the model for transitive sentences with Cartesian and semi-Cartesian verbs, using a dataset developed in \cite{KartsaklisSadrzadeh2014}. This dataset consists of 108 pairs of transitive sentences, together with human judgements on their similarity. The human judgements range from 1 to 7, and there are approximately 24 human judgements per pair of sentences. We take the mean value of the human judgements for each pair, giving us a ranking of the similarities of the pairs. The performance of each method is tested using Spearman's $\rho$, which gives a measure of the similarity of two rankings.

\subsection{Models}
We use GloVe vectors \cite{pennington2014} in order to compare our methods' efficacy on off-the-shelf pretrained word vectors, hence avoiding costly training time. We generate sentence vectors for each sentence in the dataset and calculate their similarity using the cosine similarity metric. 	

Sentence vectors were generated as follows. Given a sentence $s = subject \; verb \; object$, we have at our disposal pretrained vectors $\vect{subj}, \vect{verb}, \vect{obj} \in N$ and we wish to form a representation of $\vect{s}$, which may or may not live in $N$. We use the following methods:

\begin{itemize}
\item \textbf{Cartesian verbs} According to definition \ref{def:Cartesian}, using Cartesian verbs gives us the following sentence representation:
\[
\vect{s} = \vect{subj}\otimes \vect{verb}\otimes \vect{obj} \in N \otimes N \otimes N
\]
\item\textbf{Semi-Cartesian verbs} This combination method is given in definition \ref{def:semicart}. Since we only have at our disposal $\vect{verb}$, we represent semi-Cartesian transitive sentences as:
\[
\vect{s} = \vect{subj}\odot \vect{verb} \otimes \vect{verb}\otimes \vect{verb} \odot \vect{obj} \in N \otimes N \otimes N
\]
\item \textbf{Verb only} $\vect{s} = \vect{verb} \in N$
\item \textbf{Additive} $\vect{s} = \vect{subj} + \vect{verb} + \vect{obj} \in N$
\item\textbf{Multiplicative} $\vect{s} = \vect{subj} \odot \vect{verb} \odot \vect{obj} \in N$
\item\textbf{KS Frobenius Additive} $\vect{s} = \vect{subj} \odot \vect{verb_s} + \vect{verb_o}\odot \vect{obj} \in N$
\item \textbf{KS Frobenius Tensored} $\vect{s} = \vect{subj} \odot \vect{verb_s} \otimes \vect{verb_o}\odot \vect{obj} \in N \otimes N$
\end{itemize}

We compare the first two methods, \textbf{Cartesian verbs} and \textbf{Semi-Cartesian verbs} with three baseline methods: \textbf{Verb only}, \textbf{Additive}, and \textbf{Multiplicative}. We also report some results from \cite{KartsaklisSadrzadeh2014} for comparison, \textbf{KS Frobenius Additive} and \textbf{KS Frobenius Tensored}. These latter two results are not strictly comparable, since they use a different set of vectors, built via co-occurrence counts rather than the predictive methods of \cite{pennington2014}. Furthermore, these methods build different vectors for the subject part of the verb and for the object part of the verb.

\subsection{Results}
On this dataset, taken from \cite{KartsaklisSadrzadeh2014}, the baseline model \textbf{Additive} performs very well, giving a value higher than the inter-annotator agreement. This is a measure of the extent to which humans agreed with each other in this task, and therefore comprises a theoretical maximum. The \textbf{Cartesian verbs} model also performs highly, outperforming the \textbf{Verb only} baseline and the inter-annotator agreement. The \textbf{Semi-Cartesian verbs} model does not do so well, and this points to the fact that we will likely need to train verb vectors that include the subject and object components of the verb as well as the holistic verb meaning. 
\begin{table}[htbp]
\begin{center}
  \begin{tabular}{ l | c  c } 
    \hline
     Model & $\rho$ & $p$ \\ \hline \hline
    Verb only & 0.443 & $< 0.001$ \\ 
    Additive & 0.699 &  $< 0.001$\\
    Multiplicative  & 0.189 & $< 0.05$ \\ \hline 
    Cartesian Verbs & 0.663 & $< 0.001$ \\ 
    Semi-Cartesian Verbs & 0.389 & $< 0.001$ \\\hline  
    KS Additive & 0.428 & - \\
    KS Tensored & 0.332 & - \\ \hline
    Inter-annotator agreement & \multicolumn{2}{|c}{0.6}\\
    \hline
  \end{tabular}
\end{center}
\caption{Results for the sentence similarity task. Results are given in terms of Spearman's $\rho$, together with $p$-values where available.}
\end{table}

\section{Outlook}
We introduced the notions of Cartesian and semi-Cartesian verbs and derived a wiring for prepositions, each of which both simplify the diagrammatic representations of sentences, and significantly reduce the size of the required models. We provided examples for how these constructions work, implemented in a model of conceptual spaces, and initial experimental evidence of the validity of these constructions. We also investigated interaction between relative pronouns and the internal structure of Cartesian verbs and sentence spaces.

Evidently there is scope for further empirical work, and application to other models. Of particular interest would be the development of `visual models' that directly represent elements of our visual perception, since many prepositions naturally inhabit the visual realm \cite{gardenfors}.

\bibliographystyle{eptcs.bst}
\providecommand{\doi}[1]{\textsc{doi}:
\href{http://dx.doi.org/#1}{\nolinkurl{#1}}}
\bibliography{main}

\end{document}